%% file: arxiv.tex
\documentclass{article}

\usepackage[margin=1in]{geometry}  % This is specific to the arxiv version
\usepackage{mathpazo}
\usepackage[backend=bibtex,style=alphabetic,natbib=true]{biblatex}
\input{math_commands.tex}

\usepackage{booktabs}

\usepackage{url}
\usepackage{amsmath,amssymb,amsthm}
\usepackage{bbm}
\usepackage{xspace}
\usepackage{graphicx}
\usepackage{color}
\usepackage[colorlinks=true,linkcolor=blue,citecolor=blue,urlcolor=blue]{hyperref}
\usepackage{subcaption}
\usepackage{comment}
\usepackage{mathtools}
\usepackage{multirow}
\usepackage{threeparttable}
\usepackage{scrextend}
\usepackage{listings}
\usepackage{algorithm,algpseudocode}
\usepackage[T1]{fontenc}
\usepackage{xcolor}
\usepackage{soul}

\newcommand{\xin}{\mathbf{x}}

\title{Certified Patch Robustness via Smoothed \\ Vision Transformers}
\author{%
  Hadi Salman\thanks{Equal contribution.} \\
  MIT\\
  \texttt{hady@mit.edu} \\
  \and
  Saachi Jain\footnotemark[1] \\
  MIT \\
  \texttt{saachij@mit.edu} \\
  \and
  Eric Wong\footnotemark[1] \\
  MIT \\
  \texttt{wongeric@mit.edu} \\
  \and
  Aleksander M\k{a}dry \\
  MIT \\
  \texttt{madry@mit.edu}
}

\date{}

\begin{document}
    \maketitle
    \begin{abstract}
\input{sections/abstract}
    \end{abstract}

    \section{Introduction}
    \label{sec:intro}
    \input{sections/intro}
    \input{sections/background}

    \section{Improving certified and standard accuracies with ViTs}
    \label{sec:improve-accuracy}

\input{sections/accuracy}

    \section{Faster inference with ViTs}
    \label{sec:improve-speed}
    \input{sections/speed}

    % \section{Vision transformers and image ablations}
    % \label{sec:why}
    % \input{sections/why-transformers}
    
    % \section{Vision transformers improve derandomized smoothing}
    % \eric{we need a better section header that punches better}
    % \label{sec:improve}
    % \input{sections/improvements.tex}

    % \section{Exploring the landscape of smoothed vision transformers}
    % \label{sec:probe}
    % \input{sections/studies.tex}

    \section{Related work}
    \input{sections/related}
    
    \section{Conclusion}
    \input{sections/conclusion}

    \section{Acknowledgements}
    \input{sections/acknowledgement}

    \clearpage
    % \bibliography{bibliography/bib.bib}
    \small
    \printbibliography

    \clearpage
    % \input{sections/checklist}

    % \clearpage

    \appendix

    %\section{Cropping instead of masking for convolutional architectures}
    %\label{app:crop}

    \section{Experimental setup}
    \label{app:setup}
    \input{sections/app-setup}
    \newpage
    
    \section{Ablation sweeps}
    \label{app:ablations}
    \input{sections/app-traintime-abl.tex}
    \newpage

    \section{Dropping tokens for ViTs}
    \label{app:droptokens}
    \input{sections/app-drop-tokens.tex}
    \newpage

    \section{Strided ablations}
	\label{app:stride}
    \input{sections/app-stride.tex}
    \newpage

    \section{Block smoothing}
    \label{app:block-smoothing}
    \input{sections/app-block.tex}

    \newpage

    \clearpage
    \section{Extended experimental results}
    \label{app:tables}
    \input{sections/tables.tex}

    \newpage
    %\label{app:missingness}
    %\input{sections/app-missingness}

    %\section{Probing smoothed ViTs additional results}
    %\label{app:probing}
    %\input{sections/probing}

    %\section{Smoothing with pretrained models}
    %\label{app:pretrained}
    %\input{sections/app-pretrained.tex}

    %\section{Dropping tokens for smoothed transformers}

    \end{document}

%% file: math_commands.tex
\usepackage{amsmath,amsfonts,bm}

\def\eqref#1{equation~\ref{#1}}
\def\1{\bm{1}}

\DeclareMathAlphabet{\mathsfit}{\encodingdefault}{\sfdefault}{m}{sl}
\SetMathAlphabet{\mathsfit}{bold}{\encodingdefault}{\sfdefault}{bx}{n}

\DeclareMathOperator*{\argmax}{arg\,max}

%% file: sections/abstract.tex
Certified patch defenses can guarantee robustness of an image classifier to arbitrary changes within a bounded contiguous region. But, currently, this robustness comes at a cost of degraded standard accuracies and slower inference times. We demonstrate how using vision transformers enables significantly better certified patch robustness that is also more computationally efficient and does not incur a substantial drop in standard accuracy. These improvements stem from the inherent ability of the vision transformer to gracefully handle largely masked images.\footnote{Our code is available at \url{https://github.com/MadryLab/smoothed-vit}.}
% available at \url{https://github.mit.edu/hady/certified-vit}.}

%% file: sections/intro.tex
High-stakes scenarios warrant the development of certifiably robust models that are \emph{guaranteed} to be robust to a set of transformations. These techniques are beginning to find applications in real-world settings, such as verifying that aircraft controllers behave safely in the presence of approaching airplanes \citep{julian2019guaranteeing}, and ensuring the stability of automotive systems to sensor noise \citep{wong2020neural}. 
%However, certified robustness often comes at a steep price: computation and accuracy are significantly worse than standard models \citep{li2020sok}, while the robustness guarantees fall short of empirical approaches \citep{croce2020robustbench}. 

% In this paper, we focus specifically on models which are certifiably robust to adversarial patches. 
%---arbitrary changes within a small, contiguous region \citep{}. 
We study robustness in the context of adversarial patches---a broad class of arbitrary changes contained within a small, contiguous region. 
Adversarial patches capture the essence of a range of maliciously designed physical objects such as adversarial glasses \citep{sharif2016accessorize}, stickers/graffiti \citep{evtimov2018robust}, and clothing \citep{wu2020making}. 
Researchers have used adversarial patches to fool image classifiers \citep{brown2018adversarial}, manipulate object detectors \citep{lee2019physical, hoory2020dynamic}, and disrupt optical flow estimation \citep{ranjan2019attacking}. 
% Several empirical approaches to improve robustness to adversarial patches \citep{bafna2018thwarting, hayes2018visible, naseer2019local} have been broken \citep{tramer2020adaptive, chiang2020certified}, leading to the development of certified defenses \citep{chiang2020certified}. 

Adversarial patch defenses can be tricky to evaluate---recent work broke several empirical defenses \citep{bafna2018thwarting, hayes2018visible, naseer2019local} with stronger adaptive attacks \citep{tramer2020adaptive, chiang2020certified}. 
% Recent work \citep{tramer2020adaptive, chiang2020certified} has shown that adaptive attacks can easily break several empirical defenses to adversarial patches \citep{bafna2018thwarting, hayes2018visible, naseer2019local}. 
This motivated \emph{certified} defenses, which deliver provably robust models without having to rely on an empirical evaluation. 
However, certified guarantees tend to be modest and come at a cost: poor standard accuracy and slower inference times \citep{levine2020robustness,levine2020randomized,zhang2020clipped,xiang2021patchguard}. 
For example, a top-performing, recently proposed method reduces standard accuracy by 30\% and increases inference time by two orders of magnitude, while certifying only 13.9\% robust accuracy on ImageNet against patches that take up 2\% of the image \citep{levine2020randomized}. 
These drawbacks are commonly accepted as the cost of certification, but severely limit the applicability of certified defenses. Does certified robustness really need to come at such a high price? 

\subsection*{Our contributions}
In this paper, we demonstrate how to leverage vision transformers (ViTs) \citep{dosovitskiy2020image} to create certified patch defenses that achieve significantly higher robustness guarantees than prior work. 
Moreover, we show that certified patch defenses with ViTs can actually maintain standard accuracy and inference times comparable to standard (non-robust) models. 
At its core, our methodology exploits the token-based nature of attention modules used in ViTs to gracefully handle the ablated images used in certified patch defenses. 
Specifically, we demonstrate the following: 
% In this paper, we demonstrate how vision transformers enable certified patch defenses that can actually maintain standard accuracy and inference times comparable to standard (non-robust) models---while achieving higher robustness guarantees than prior work.  Our methodology exploits the token-based, attention modules used in vision transformers \citep{dosovitskiy2020image, wu2020visual} to gracefully handle highly masked images commonly used in certified patch defenses. 
% In particular, we demonstrate the following: 

% In this paper, we show that typical robustness/performance trade-offs associated with certified robustness are not as severe when it comes to adversarial patches. 
% In this paper, we show how typical robustness/performance trade-offs associated with certified robustness can be significantly alleviated using a more suitable architecture. 
% In this paper, we demonstrate how architectures more suited for patch defenses can significantly alleviate typical robustness/performance trade-offs associated with certified robustness. 
% \eric{need to think about rephrasing this first sentence to avoid passive voice}
% Specifically, we find that using Vision Transformers \citep{dosovitskiy2020image, wu2020visual} as the backbone for a certified patch defense maintains high standard accuracy, reduce computational complexity, and enable substantial gains in certified patch performance. We concretely show the following: 

\paragraph{Improved guarantees via smoothed vision transformers.}
We find that using ViTs as the backbone of the derandomized smoothing defense \citep{levine2020randomized} enables significantly improved certified patch robustness. Indeed, this change alone boosts certified accuracy by up to 13\% on ImageNet, and 5\% on CIFAR-10 over similarly sized ResNets. 

\paragraph{Standard accuracy comparable to that of standard architecures.}
We demonstrate that ViTs enable certified defenses with standard accuracies comparable to that of standard, non-robust models. 
In particular, our largest ViT improves state-of-the-art certified robustness on ImageNet while maintaining standard accuracy that is similar to that of a non-robust ResNet (>70\%).

\paragraph{Faster inference.}
% \eric{Maybe someone can take a stab at adjusting this paragraph header or does it look ok? might depend on how close we can actually get to standard inference speeds}
We modify the ViT architecture to drop unnecessary tokens, and reduce the smoothing process to pass over mostly redundant computation. 
These changes turn out to vastly speed up inference time for our smoothed ViTs. 
In our framework, a forward pass on ImageNet becomes up to two orders of magnitude faster than that of prior certified defenses, and is close in speed to a standard (non-robust) ResNet. 

%% file: sections/background.tex
\section{Certified patch defense with smoothing \& transformers}
Smoothing methods are a general class of certified defenses that combine the predictions of a classifier over many variations of an input to create predictions that are certifiably robust \citep{cohen2019certified, levine2020robustness}. 
One such method that obtains robustness to adversarial patches is derandomized smoothing \citep{levine2020randomized}, which aggregates a classifier's predictions on various \textit{image ablations} that mask most of the image out.

These approaches typically use CNNs, a common default model for computer vision tasks, to evaluate the image ablations. The starting point of our approach is to ask: are convolutional architectures the right tool for this task? The crux of our methodology is to leverage vision transformers, which we demonstrate are more capable of gracefully handling the image ablations that arise in derandomized smoothing. 

\subsection{Preliminaries}
\label{sec:preliminaries}
\begin{figure}[!htbp]
    \centering
        \includegraphics[width=1\textwidth]{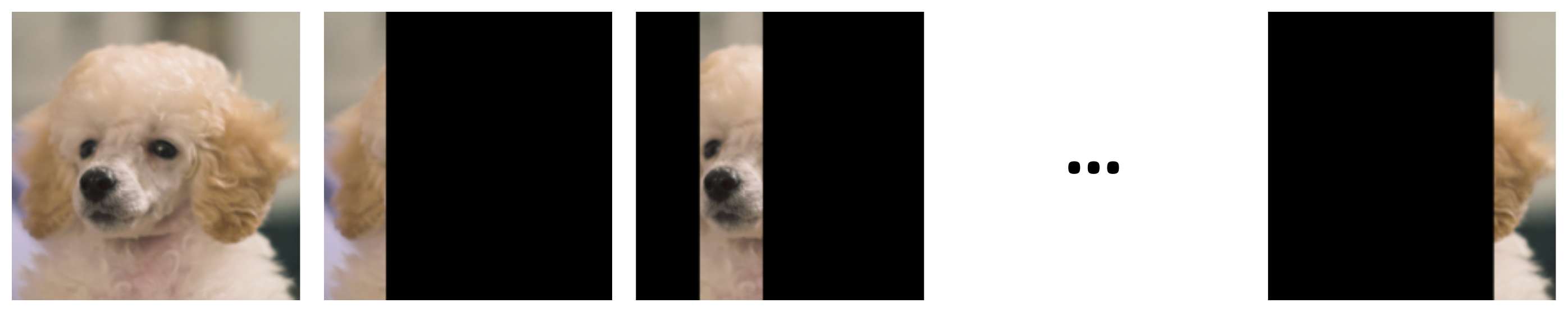}
        \caption{Examples of column ablations for the left-most image with column width 19px.}
    \label{fig:example-ablations}
\end{figure}

\paragraph{Image ablations.}
Image ablations are variations of an image where all but a small portion of the image is masked out \citep{levine2020randomized}. For example, a column ablation masks the entire image except for a column of a fixed width (see Figure \ref{fig:example-ablations} for an example). We focus primarily on column ablations and explore the more general block ablation in Appendix \ref{app:block-smoothing}. 
% \textit{Image ablations}, where all but a small portion of the input image is masked out, are a crucial part of de-randomized smoothing. Masking is performed by blacking out the pixels that need to be hidden. In this paper, we focus on \textit{column} ablations, which mask the entire image except for a column of width $b$ \cite{levine2020randomized}. This column can wrap around the edge of the image. We show examples of column ablations in Figure \ref{fig:example-ablations}.

For a input $h \times w$ sized image $\mathbf{x}$, we denote by $\mathcal S_b(\mathbf{x})$ the set of all possible column ablations of width $b$. 
A column ablation can start at any position and wrap around the image, so there are $w$ total ablations in $\mathcal S_b(\mathbf{x})$.

% While we focus on column ablations in this paper, \textit{block ablations} (where everything except a square block is masked) are another common mechanism for generating image ablations. We consider block ablations in Appendix \ref{app:block-smoothing} \saachi{is this the right place to put this reference?}.

% \subsection{Building a smoothed classifier}
% \subsection{Preliminaries: derandomized smoothing}

\begin{figure}
    \includegraphics[width=\textwidth]{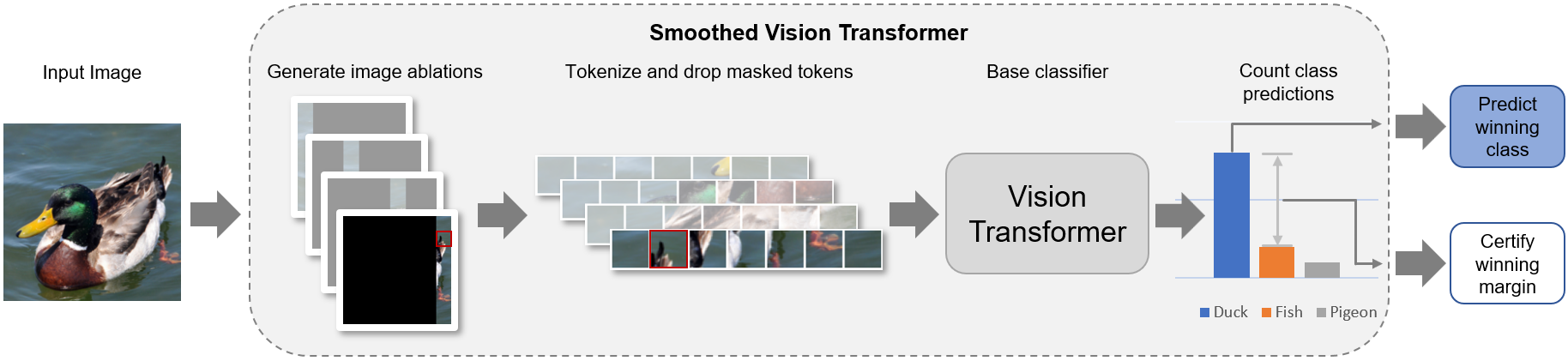}
    \caption{Illustration of the smoothed vision transformer. For a given image, we first generate a set of ablations. We encode each ablation into tokens, and drop fully masked tokens. The remaining tokens for each ablation are then fed into a vision transformer, which predicts a class label for each ablation. We predict the class with the most predictions over all the ablations, and use the margin to the second-place class for robustness certification.}
    \label{fig:overview}
\end{figure}

\paragraph{Derandomized smoothing.}
%Now that we have specified our set of image ablations, we are ready to define \textit{derandomized smoothing} \cite{levine2020randomized}. 
% let $f$ be the \textit{base classifier} which, given an image ablation such as those in Figure \ref{fig:example-ablations}, outputs a class label. 
Derandomized smoothing \citep{levine2020randomized} is a popular approach for certified patch defenses that constructs a \textit{smoothed classifier} comprising of two main components: (1) a \emph{base classifier}, and (2) a set of image ablations used to smooth the base classifier. 
% Let $f$ be the base classifier that outputs a class label when given an image. 
Then, the resulting smoothed classifier returns the most frequent prediction of the base classifier over the ablation set $\mathcal{S}_b(\mathbf{x})$. Specifically, for an input image $\xin$, ablation set $\mathcal S_b(\xin)$, and a base classifier $f$, a smoothed classifier $g$ is defined as: 
\begin{equation}
g(\xin) = \argmax_{c} n_c(\xin)
\label{eq:smooth}
\end{equation}
where 
$$n_c(\xin) = \sum_{\xin' \in S_b(\xin)} \mathbb{I}\{f(\xin') = c\}$$
denotes the number of image ablations that were classified as class $c$. We refer to the fraction of images that the smoothed classifier correctly classifies as \emph{standard accuracy}.

% For an input image $x$, we pass every ablation of that image, $x' \in S_b(x)$ to the base classifier. Let $n_c$ thus be the \textit{number} of image ablations which were classified as class $c$:
% $$n_c(\mathbf{x}) = \sum_{x' \in S_b(\mathbf{x})} \mathbb{I}\{f(x') = c\}.$$
% Then derandomized smoothing constructs a new \textit{smoothed classifier} $g$ which aggregates the predictions of the base classifier on all of these ablations and outputs the class that was predicted most often:
% $$g(\mathbf{x}) = \argmax_{c} n_c(\mathbf{x}).$$

% We call the accuracy of the smoothed classifier $g$ the \textit{standard accuracy}.

% \subsection{Certification using derandomized smoothing.}
A smoothed classifier is \textit{certifiably robust} for an input image if the number of ablations for the most frequent class exceeds the second most frequent class by a large enough margin. 
Intuitively, a large margin makes it impossible for an adversarial patch to change the prediction of a smoothed classifier since a patch can only affect a limited number of ablations. 

Specifically, let $\Delta$ be the maximum number of ablations in the ablation set $S_b(\mathbf{x})$ that an adversarial patch can simultaneously intersect (e.g., for column ablations of size $b$, an $m \times m$ patch can intersect with at most $\Delta = m + b - 1$ ablations). 
Then, a smoothed classifier is certifiably robust on an input $\xin$ if it is the case that for the predicted class $c$: 
\begin{equation}
n_c(\mathbf{x}) > \max_{c' \neq c} n_{c'}(\mathbf{x}) + 2\Delta.
\label{eq:certify}
\end{equation}
If this threshold is met, the most frequent class is guaranteed to not change even if an adversarial patch compromises every ablation it intersects. 
We denote the fraction of predictions by the smooth classifier that are both correct and certifiably robust (according to Equation \ref{eq:certify}) as \textit{certified accuracy}. 

\paragraph{Vision transformers.}
A key component of our approach is the vision transformer (ViT) architecture  \citep{dosovitskiy2020image}. In contrast to convolutional architecures, ViTs use self-attention layers instead of convolutional layers as their primary building block and are inspired by the success of self-attention in natural language processing \citep{vaswani2017attention}. 
ViTs process images in three main stages: %tokenization and self-attention.
% The performance and guarantees of smoothed models hinge on the ability of the base classifier to quickly and accurately classify image ablations. 
% Although the base classifier could be any black box model, they have thus far largely been studied in the context of convolutional neural networks (CNNs). 

% In this paper, we show that vision transformers are better equipped than CNNs for the setting of derandomized smoothing. Proposed by \citet{dosovitskiy2020image}, vision transformers (ViTs) use self-attention instead of convolutions to evaluate images. ViTs proceed in two main stages: tokenization and self-attention.
\begin{enumerate}
\item \textit{Tokenization:} The ViTs split the image into $p \times p$ patches. Each patch is then embedded into a positionally encoded \textit{token}.
\item \textit{Self-Attention:} The set of tokens are then passed through a series of multi-headed self-attention layers \citep{vaswani2017attention}. % followed by a classification head of fully connected layers
\item \textit{Classification head:} The resulting representation is fed into a fully connected layer to make predictions for classification. 
\end{enumerate}
% In contrast to CNNs, which gradually increase their receptive field with additional layers, ViTs \textit{globally} share information across the image at every layer \citep{vaswani2017attention}.

\begin{table}[t]
    \begin{minipage}{\textwidth}
        \renewcommand\footnoterule{}
        \centering
        \caption{Summary of our ImageNet results and comparisons to certified patch defenses from the literature: Clipped Bagnet (CBG), Derandomized Smoothing (DS), and PatchGuard (PG). Time refers to the inference time for  a batch of 1024 images, $b$ is the ablation size, and $s$ is the ablation stride. An extended version is in Appendix~\ref{app:tables}.}
            \begin{tabular}{l|cccc|c}
            \toprule
            \multicolumn{6}{c}{Standard and Certified Accuracy on ImageNet (\%) }\\
            \midrule
            & Standard & 1\% pixels & 2\% pixels & 3\% pixels & Time (sec) \\
            \midrule
            \multicolumn{5}{l}{\textit{Baselines} } \\
            \midrule
            Standard ResNet-50 & 76.1 & --- & --- & --- & 0.67\\
            WRN-101-2 & 78.85 & --- & --- & --- & 3.1\\
            ViT-S & 79.90 & --- & --- & --- & 0.4\\
            ViT-B & 81.80 & --- & --- & --- & 0.95\\
            CBN~\citep{zhang2020clipped} & 49.5 & 13.4 & 7.1&  3.1 & {3.05} \\ 
            DS~\citep{levine2020randomized}\footref{fn:ds} & 44.4 &17.7 & 14.0&  11.2 & {149.5}\\ 
            PG~\citep{xiang2021patchguard}\footref{fn:pg2} & $55.1$\footref{fn:pg2} & $32.3$\footref{fn:pg2} & $26.0$\footref{fn:pg2}&  $19.7$\footref{fn:pg2}  & ${3.05}$\\
            \midrule
            \multicolumn{5}{l}{\textit{Smoothed models} } \\
            \midrule
            ResNet-50 (b = 19) & 51.5 & 22.8 & 18.3 & 15.3  & 149.5 \\
            ViT-S (b = 19) & \textbf{63.5} &    \textbf{36.8} &  \textbf{31.6} &  \textbf{27.9}  & \textbf{14.0}\\
            \midrule
            WRN-101-2 (b = 19) & 61.4 &    33.3 &  28.1 &  24.1 & 694.5 \\
            ViT-B (b = 19) & 69.3 & \textbf{43.8} & \textbf{38.3}   & \textbf{34.3} & 31.5 \\          
            ViT-B (b = 37) & \textbf{73.2} &    43.0 &  38.2 &  34.1 & 58.7 \\
            ViT-B (b = 19, s = 10) &  68.3 &  36.9 &    36.9&  31.4& \textbf{3.2}\\
            \bottomrule
        \end{tabular}
        \label{table:main_summary table}
    \end{minipage}
\end{table}

\begin{figure}[b]
    \begin{minipage}{\textwidth}
        \footnotetext[1]{\label{fn:ds}We found that ResNets could achieve a significantly higher certified accuracy than was reported by \citet{levine2020randomized} if we use early stopping-based model selection. We elaborate further in Appendix \ref{app:setup}.}
        \footnotetext[2]{\label{fn:pg2}The PatchGuard defense uses a specific mask size that guarantees robustness to patches smaller than the mask, and provides no guarantees for larger patches.  In this table, we report their best results: each patch size corresponds to a separate model that achieves 0\% certified accuracy against larger patches. Comparisons across the individual models can be found in Appendix \ref{app:tables}.}
    \end{minipage}        
\end{figure}

\subsection{Smoothed vision transformers}
\label{sec:smoothed-vit}

% Our goal is to combine the certification strengths of derandomized smoothing with the token-based architecture of ViTs. Intuitively, the ViT architecture is appealing for processing image ablations (and consequently derandomized smoothing) for two main reasons: 

Two central properties of vision transformers make ViTs particularly appealing for processing the image ablations that arise in derandomized smoothing.  Firstly, unlike CNNs, ViTs process images as sets of tokens. ViTs thus have the natural capability to simply drop unnecessary tokens from the input and ``ignore'' large regions of the image, which can greatly speed up the processing of image ablations. 

Moreover, unlike convolutions which operate locally, the self-attention mechanism in ViTs shares information \emph{globally} at every layer \citep{vaswani2017attention}. Thus, one would expect ViTs to be better suited for classifying image ablations, as they can dynamically attend to the small, unmasked region. In contrast, a CNN must gradually build up its receptive field over multiple layers and process masked-out pixels.

Guided by these intuitions, our methodology leverages the ViT architecture as the base classifier for processing the image ablations used in derandomized smoothing. 
We first demonstrate that these \emph{smoothed vision transformers} enable substantially improved robustness guarantees, without losing much standard accuracy (Section \ref{sec:improve-accuracy}). 
We then modify the ViT architecture and smoothing procedure to drastically speed up the cost of inference of a smoothed ViT (Section \ref{sec:improve-speed}). 
We present an overview of our approach in Figure \ref{fig:overview}. 

% to modify the ViT architecture to drop masked tokens from input images, enabling the ViT to gracefully handle image ablations. 
% We then create a smoothed version of our ViT using Equation \ref{eq:smooth}, namely the \emph{smoothed vision transformer}, that is certifiably robust according to Equation \ref{eq:certify}.

\paragraph{Setup.}
We focus primarily on the column smoothing setting and defer block smoothing results to Appendix \ref{app:block-smoothing}. We consider the CIFAR-10~\citep{krizhevsky2009learning} and ImageNet~\citep{deng2009imagenet} datasets, and perform our analysis on three sizes of vision transformers---ViT-Tiny (ViT-T), ViT-Small (ViT-S), and ViT-Base (ViT-B) models \citep{rw2019timm, dosovitskiy2020image}.
We compare to residual networks of similar size---ResNet-18, ResNet-50 \citep{he2016deep}, and Wide ResNet-101-2 \citep{zagoruyko2016wide}, respectively.
% \footnote{Wide ResNet-101-2 is slightly larger than Vit-B.}
Further details of our experimental setup are in Appendix \ref{app:setup}.

%% file: sections/accuracy.tex
\begin{table}
    \begin{minipage}{\textwidth}
        \caption{Summary of our CIFAR-10 results and comparisons to certified patch defenses from the literature: Clipped Bagnet (CBG), Derandomized Smoothing (DS), and PatchGuard (PG). Here, $b$ is the column ablation size out of 32 pixels. An extended version is in Appendix~\ref{app:tables}.}
    \centering
    \begin{tabular}{l|ccc}
        \toprule
        \multicolumn{4}{c}{Standard and Certified Accuracy on CIFAR-10 (\%) }\\
        \midrule
         & Standard & $2\times2$ & $4\times4$ \\
        \midrule
        \multicolumn{4}{l}{\textit{Baselines} } \\
        \midrule
        CBN~\citep{zhang2020clipped} & 84.2 & 44.2 & 9.3\\ 
        DS~\citep{levine2020randomized}\footref{fn:ds} & 83.9 & 68.9 & 56.2\\ 
        PG~\citep{xiang2021patchguard}\footref{fn:pg2} & 84.7\footref{fn:pg2} &  69.2\footref{fn:pg2} & 57.7\footref{fn:pg2} \\
        % PG~\cite{xiang2021patchguard} ($4\times 4$) & 84.6 &  57.7 & 57.7 \\
        \midrule
        \multicolumn{4}{l}{\textit{Smoothed models} } \\
        % \midrule
        % ResNet-18 (b = 4) & 83.6 &   67.0 &  54.2\\
        % ViT-T (b = 4) & \textbf{85.5} &    \textbf{70.0} &  \textbf{58.5}  \\
        \midrule
        ResNet-50 (b = 4) & 86.4 &    71.6 &  59.0 \\
        ViT-S (b = 4) & \textbf{88.4} &    \textbf{75.0} &  \textbf{63.8} \\
        \midrule
        WRN-101-2 (b = 4) & 88.2 &    73.9 &  62.0 \\
        ViT-B (b = 4) & \textbf{90.8} &    \textbf{78.1} &  \textbf{67.6}  \\
        % \midrule
        \bottomrule
    \end{tabular}
    \label{table:summary table_cifar}
\end{minipage}
\end{table}

Recall that even though certified patch defenses can guarantee robustness to patch attacks, this robustness typically does not come for free. 
Indeed, certified patch defenses tend to have substantially lower standard accuracy when compared to typical (non-robust) models, while delivering a fairly limited degree of (certified) robustness.

% In addition to slow inference times, certified patch defenses with CNNs suffer from substantially degraded standard accuracy compared to standard (non-robust) models. Fundamentally, CNNs struggle to accurately label the highly masked inputs used in derandomized smoothing. This not only bottlenecks the standard accuracy of the smooth classifier but results in worse robustness guarantees. 

In this section, we show how to use ViTs to substantially improve both standard and certified accuracies for certified patch defenses. 
To this end, we first empirically demonstrate that ViTs are a more suitable architecture than traditional convolutional networks for classifying the image ablations used in derandomized smoothing (Section \ref{sec:ablations}). 
Specifically, this change in architecture alone yields models with significantly improved standard and certified accuracies. 
We then show how a careful selection of smoothing parameters can enable smoothed ViTs to have even higher standard accuracies that are comparable to typical (non-robust) models, without sacrificing much certified performance (Section \ref{sec:ablation_size}). 

Our ImageNet and CIFAR-10 results are summarized in Table~\ref{table:main_summary table} and Table~\ref{table:summary table_cifar}, respectively. We further include the inference time to evaluate a batch of images, using the modifications described in Section~\ref{sec:improve-speed}. See Appendix~\ref{app:tables} for extended tables covering a wider range of experiments.

% Finally in Section~\ref{sec:improve-speed}, we modify both the ViT architecture and the derandomized smoothing process to significantly speed up certification using smoothed ViTs. We are able to vastly improve inference time for smoothed ViTs, making them comparable in speed to standard (non-robust) convolutional architectures. A summary of certification costs for our ViTs is shown in Table~\ref{table:main_summary table}.

\subsection{ViTs outperform ResNets on image ablations.}
\label{sec:ablations}
% \eric{is there a good name for these two properties?}
% \eric{Consider moving CIFAR10 to appendix, for space}
% The success of derandomized smoothing hinges entirely on whether the base classifier can accurately classify subregions of an image. 
We first isolate the effect of using a ViT instead of a ResNet as the base classifier for derandomized smoothing. 
Specifically, we keep all smoothing parameters fixed and only vary the base classifier. Following \citet{levine2020randomized}, we use column ablations of width $b=4$ for CIFAR-10 and $b=19$ for ImageNet for both training and certification. 

\begin{figure}[!t]
    \centering
    \begin{subfigure}[b]{0.48\textwidth}
    \centering
        \includegraphics[width=\textwidth]{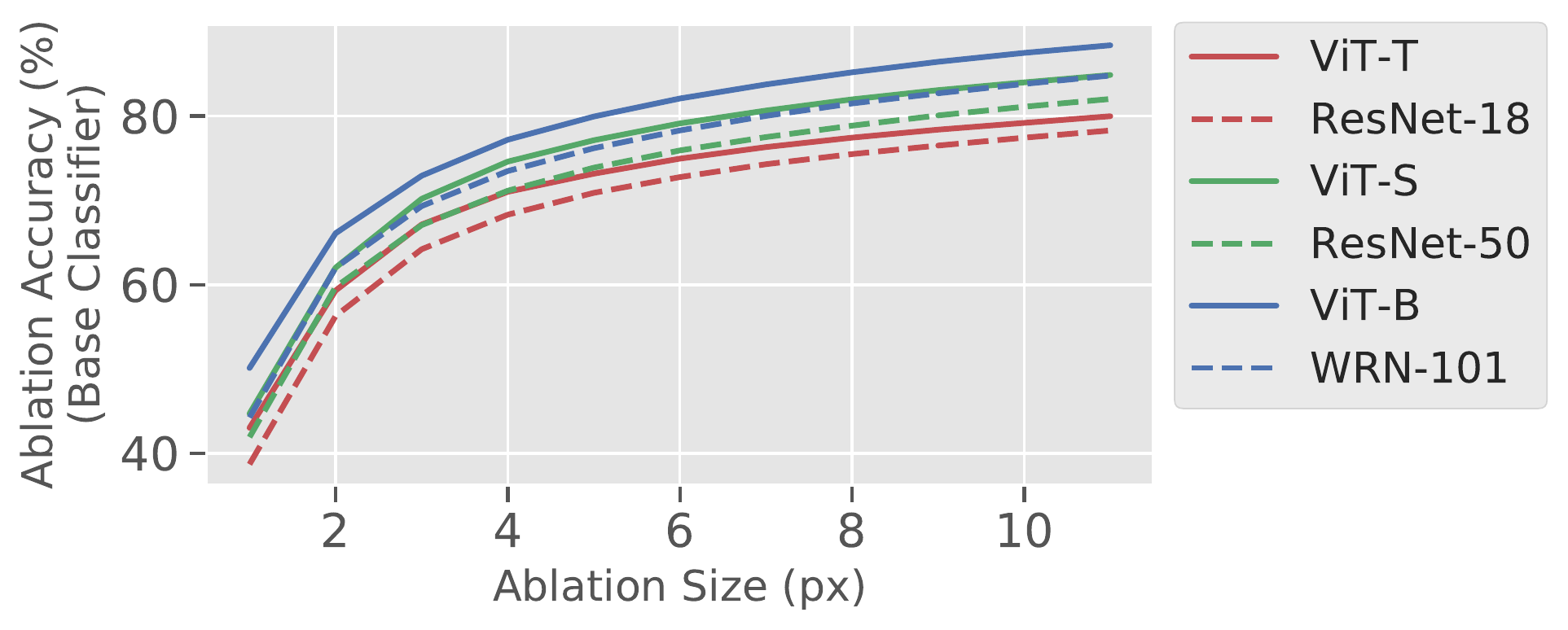}
        \caption{CIFAR-10}
    \end{subfigure}
    \begin{subfigure}[b]{0.48\textwidth}
        \centering
        \includegraphics[width=\textwidth]{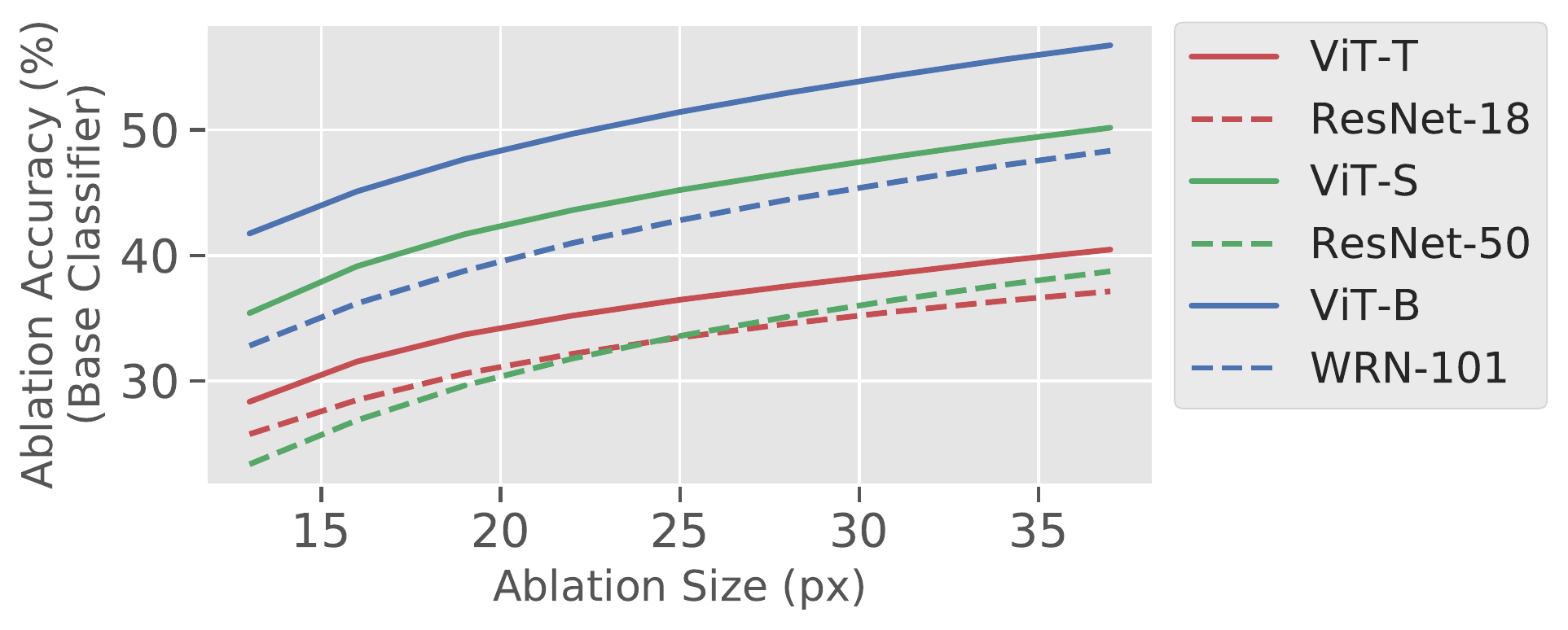}
        \caption{ImageNet}
    \end{subfigure}
\caption{Accuracies on column-ablated images for models on CIFAR-10 and ImageNet. The models were trained on column ablations of width $b=19$ for ImageNet and $b=4$ for CIFAR-10, and evaluated on a range of ablation sizes. ViTs outperform ResNets on image ablations by a sizeable margin.}
\label{fig:finetuned}
\end{figure}

\paragraph{Ablation accuracy.}
% The success of de-randomized smoothing methods hinges entirely on whether the base classifier can accurately classify subregions of an image, such as a small column or a patch. 
The performance of derandomized smoothing entirely depends on whether the base classifier can accurately classify ablated images. 
We thus measure the accuracy of ViTs and ResNets at classifying column ablated images across a range of evaluation ablation sizes as shown in Figure~\ref{fig:finetuned}. We find that ViTs are significantly more accurate on these ablations than comparably sized ResNets. 
For example, on ImageNet, ViT-S has up to 12\% higher accuracy on ablations than ResNet-50. 
% These improvements in ab

% We find that ViTs are significantly more accurate on column ablations than comparably sized ResNets.  The difference is most prominent on ImageNet---ViT-S has 11-12\% higher ablation accuracy across a range of ablation sizes than the the comparably sized ResNet-50. Since ViTs are better at classifying image ablations, we expect ViTs can improve the performance of derandomized smoothing.

\begin{figure}[!t]
    \centering
    \begin{subfigure}[b]{0.48\textwidth}
        \centering
            \includegraphics[trim=0 0 0 0, clip, width=1\textwidth]{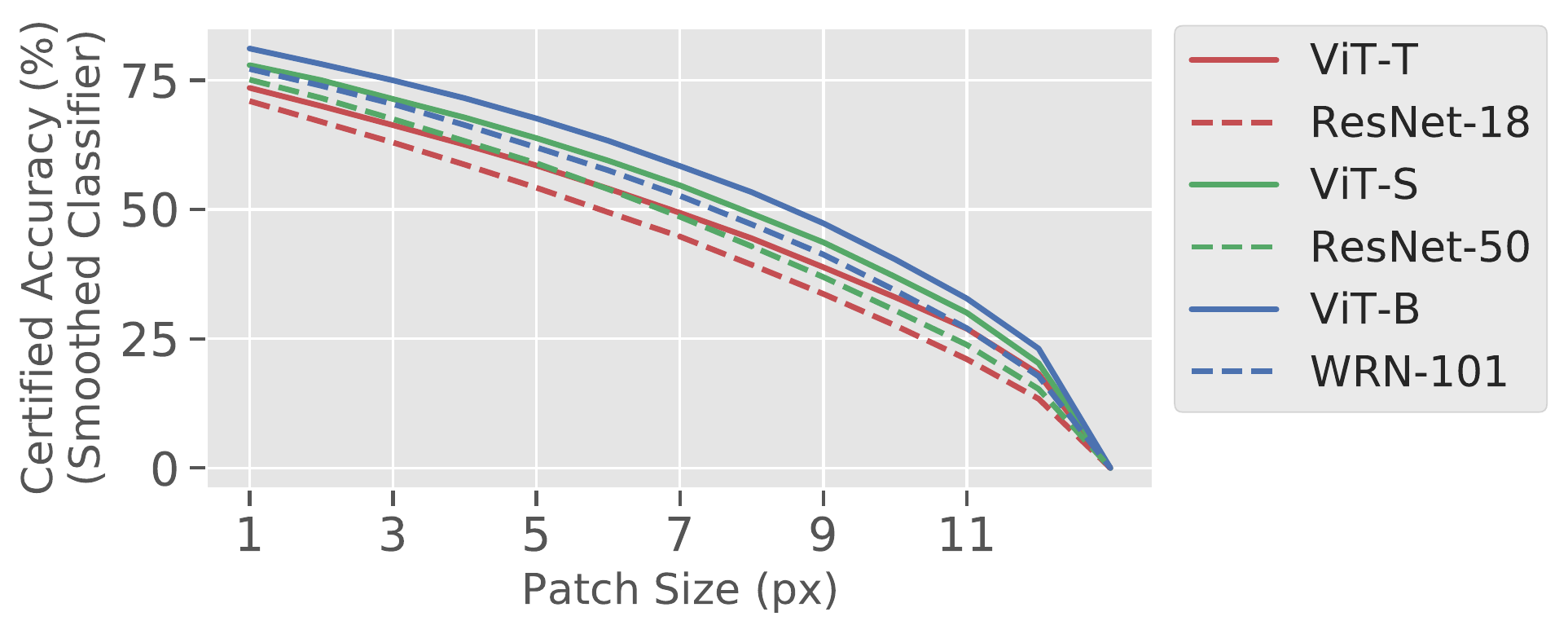}
            \caption{CIFAR-10}
        \end{subfigure}
        \begin{subfigure}[b]{0.48\textwidth}
            \centering
            \includegraphics[width=1\textwidth]{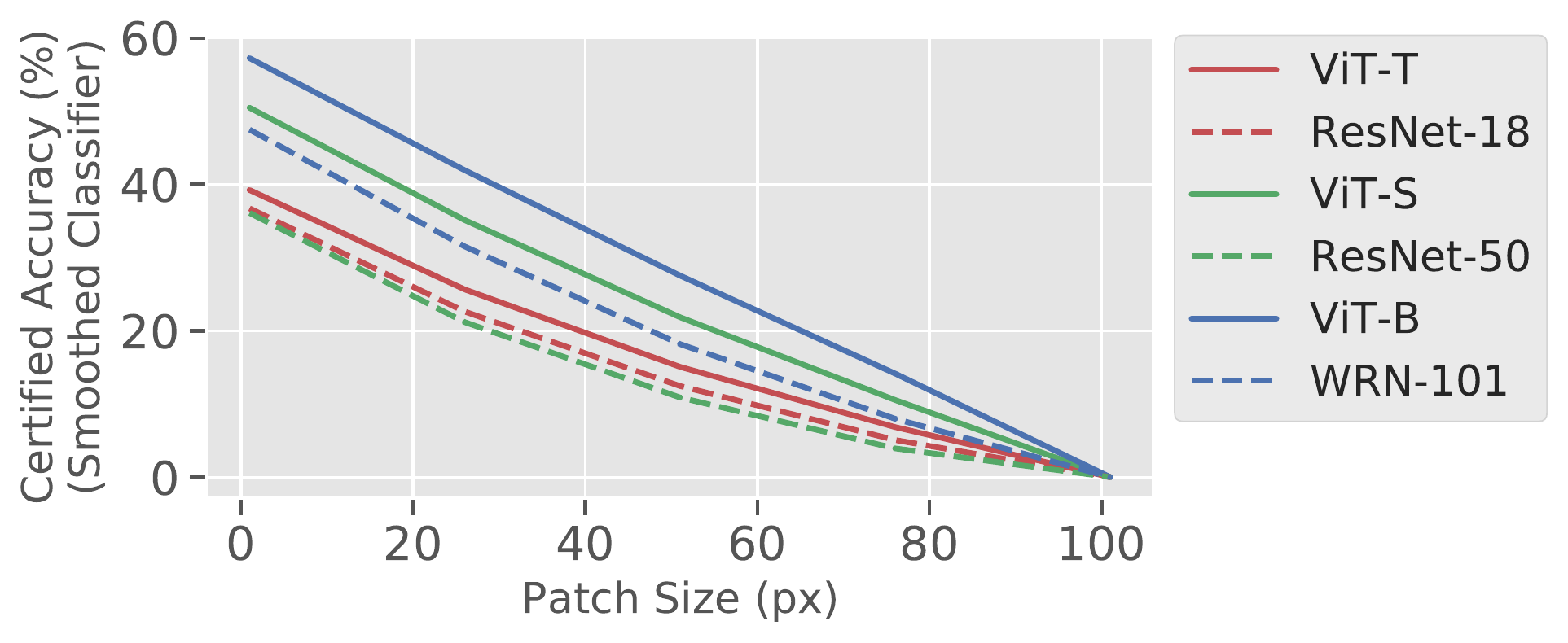}  
            \caption{ImageNet}
        \end{subfigure}
\caption{Certified accuracies for ViT and ResNet models on CIFAR-10 and ImageNet for various adversarial patch sizes. Certification was performed using a fixed ablation of size $b=4$ for CIFAR-10 and $b=19$ for ImageNet (as in \cite{levine2020randomized}).}
\label{fig:certified-vs-patchsize}
\end{figure}

% \begin{table}
%     \caption{Standard accuracies for smoothed ViTs and ResNets on CIFAR-10 and ImageNet. Certification was performed was performed using a fixed ablation size $b=4$ for CIFAR-10 and $b=19$ for ImageNet. 
%     \ericinline{delineate with horizontal rules instead of vertical}}
%     \centering
%     \begin{tabular}{lcccccccc}
%         \toprule
%         {} & \multicolumn{8}{c}{Standard Accuracy of Smoothed Classifier (\%)} \\
%         Architecture &             ViT-T &   ResNet-18 && ViT-S &  ResNet-50 && ViT-B & WRN-101 \\
%         % \midrule
%         \cmidrule{2-3} \cmidrule{5-6} \cmidrule{8-9}
%         ImageNet &            \textbf{52.25} &  50.62 &&    \textbf{63.48} &  51.47 &&   \textbf{69.33} &     61.38 \\
%         CIFAR-10 &            \textbf{86.03} &  84.75 &&    \textbf{89.36} &  87.42 &&   \textbf{91.73} &     89.10 \\
%                 \bottomrule
%     \end{tabular}
%     \label{table:standard-smoothed-acc}
% \end{table}

\paragraph{Certified patch robustness.} 
% Certification in derandomized smoothing depends on whether the base classifier correctly predicts enough image ablations \ref{eq:certify}. 
% Higher ablation accuracies for the base classifier result in smoothed classifiers with better certified accuracy. 
% Since certification of patch robustness depends the base classifier predicting the correct class over a large enough margin, any improvement in the base classifier for image ablations leads directly to better certified accuracy. 
We next measure the effect of improved ablation accuracy on certified accuracy. 
We find that using a ViT as the base classifier in derandomized smoothing substantially boosts certified accuracy compared to ResNets across a range of model sizes and adversarial patch sizes, as shown in Figure~\ref{fig:certified-vs-patchsize}. For example, against $32\times 32$ adversarial patches on ImageNet (2\% of the image), a smoothed ViT-S improves certified accuracy by 14\% over a smoothed ResNet-50, while the larger ViT-B reaches a certified accuracy of 39\%---well above the highest reported baseline of 26\% \citep{xiang2021patchguard}\footnote{The highest reported certified accuracy in the literature for this patch size on ImageNet is 26\% from PatchGuard \citep{xiang2021patchguard}. However, this defense uses a masking technique that is optimized for this particular patch size, and achieves 0\% certified accuracy against larger patches.}.
% \eric{Need to include the baselines in either a table or a figure}

\paragraph{Standard accuracy.}
We further find that smoothed ViTs can mitigate the precipitous drop in standard accuracy observed in previously proposed certified defenses, particularly so for larger architectures and datasets. 
Indeed, the smoothed ViT-B remains 69\% accurate on ImageNet---14.2\% higher standard accuracy than that of the best performing prior work (Table \ref{table:main_summary table}). 
A full comparison between the performance of smoothed models and their non-robust counterparts can be found in Appendix \ref{app:tables}.
% We further find that using ViTs can help counter the precipitous drop in standard accuracy often observed in certifiable robust models. A smoothed ResNet-50 drops over 27\% in standard accuracy over a regular ResNet-50. However, we find that \svit{} do not suffer from as dramatic of a degradation. For example, the smoothed ViT-B has a standard accuracy of 69\%, which is 13\% lower than a standard ViT-B, as shown in Table~\ref{table:standard-smoothed-acc}. 
% \saachi{is it weird that we are using resnet50 and vit-b here, which are different sizes.}
% \eric{I agree this is weird}

\subsection{Ablation size matters}
\label{sec:ablation_size}
In the previous section, we fixed the width of column ablations at $b=19$ for derandomized smoothing on ImageNet, following \cite{levine2020randomized}. 
%However, derandomized smoothing depends heavily on the \textit{ablation size}; for column smoothing, this is the width of the column of pixels that is not masked. Larger ablation sizes can make each image ablation easier to classify, but also increases the necessary threshold for certification (see Equation~\ref{eq:certify}). 
We now demonstrate that properly choosing the ablation size can improve the standard accuracy even further---by 4\% on ImageNet---without sacrificing certified performance.

% Previously in section~\ref{sec:ablations}, we saw that base classifiers can more accurately classify larger ablation sizes, but how does this affect certified robustness? 
% After all, larger ablations directly increase the necessary threshold for certification and reduce certified performance (as shown in Equation~\ref{eq:certificate}). In this section, we find that certified performance is largely stable with respect to ablation size, allowing for a trade-off between computation time and clean accuracy without losing much certified performance, particularly on ImageNet. 

\begin{figure}[!htbp]
    \centering
    % \begin{subfigure}[b]{.8\textwidth}
    % \centering
    %     \includegraphics[width=.48\textwidth]{figures/Finetuned/cifar10_cert_acc.pdf}
    %     \includegraphics[width=.48\textwidth]{figures/Finetuned/cifar10_smooth_acc.pdf}
    %     \caption{CIFAR-10 models.}
    % \end{subfigure}
    \begin{subfigure}[b]{1\textwidth}
        \centering
            \includegraphics[width=.49\textwidth]{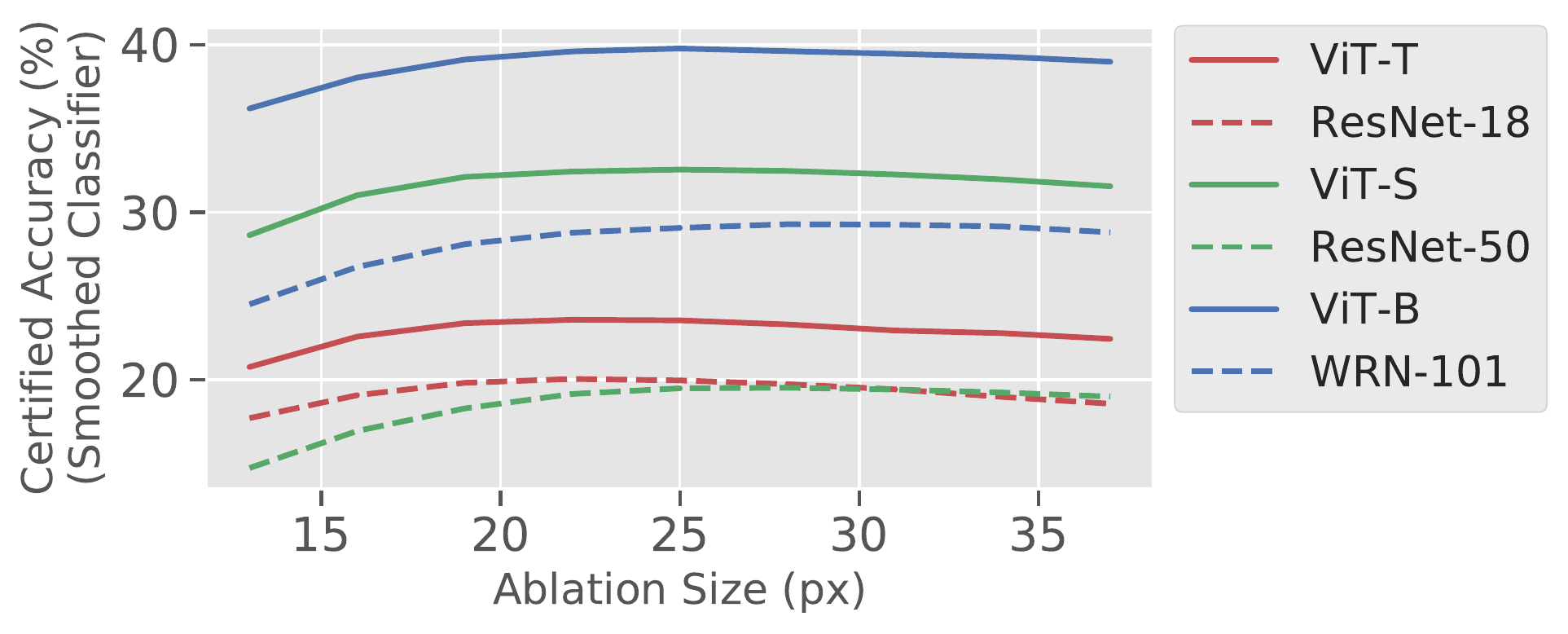}
            \includegraphics[width=.49\textwidth]{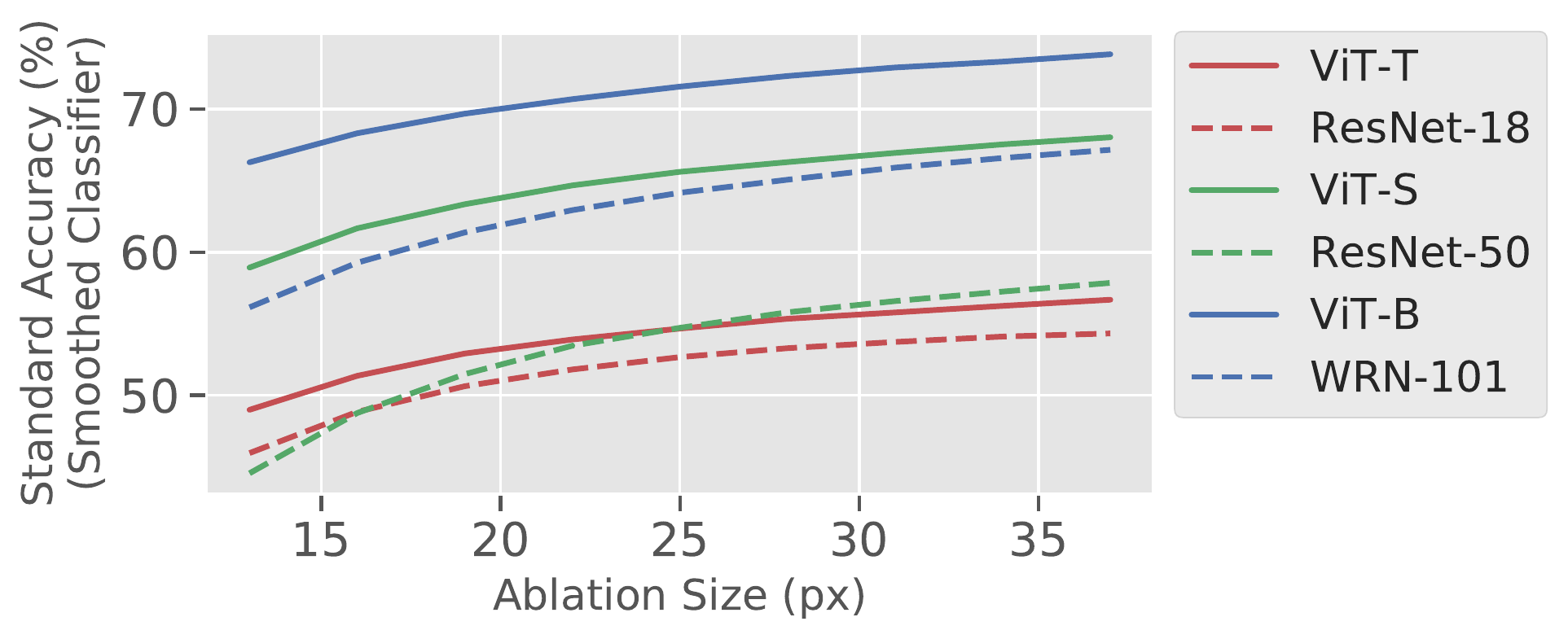}
            % \caption{ImageNet models.}
        \end{subfigure}
    \caption{
    Certified (left) and standard (right) accuracies for a collection of smoothed models trained with a fixed ablation size $b=19$ on ImageNet, and evaluated with varying ablation sizes. Certified accuracy remains stable across a range of ablation sizes, while standard accuracy substantially improves with larger ablations. 
    % Certified and standard accuracy for a smoothed model trained with a fixed ablation size ($b=5$ for CIFAR-10 and $b=19$ for ImageNet), and evaluated with varying ablation sizes. For ImageNet, certified accuracy remains stable across a range of ablation sizes, while clean accuracy can substantially improve with larger ablations. 
    }
    \label{fig:ablation-size-finetuned}
\end{figure}

% focusing just on ImageNet
Specifically, we take ImageNet models trained on column ablations with width $b=19$, 
and change the smoothing procedure to use a different width at \emph{test} time.  
We report the resulting standard and certified accuracies in Figure~\ref{fig:ablation-size-finetuned}, and defer additional experiments on changing the ablation size during training to Appendix~\ref{app:trainablations}.

Although \citet{levine2020randomized} found a steep trade-off between certified and standard accuracy in CIFAR-10 (which we verify in Appendix~\ref{app:testablations}), we find this to not be the case for ImageNet for either CNNs or ViTs. 
We can thus substantially increase the ablation size to improve standard accuracy \emph{without} significantly dropping certified performance as shown in Figure~\ref{fig:ablation-size-finetuned}. 
For example, increasing the width of column ablations to $b=37$ improves the standard accuracy of the smoothed ViT-B model by nearly $4\%$ to 73\% while maintaining a 38\% certified accuracy against $32\times 32$ patches. In addition to being 12\% higher than the standard accuracy of the best performing prior work, this model's standard accuracy is only 3\% lower than that of a \textit{non-robust} ResNet-50. 

Thus, using smoothed ViTs, we can achieve state-of-the-art certified robustness to patch attacks in the ImageNet setting while attaining standard accuracies that are more comparable to those of non-robust ResNets. 
% Therefore, in combination with the the fast inference modification of ViTs that we next present in Section \ref{sec:improve-speed}, our smoothed ViT-B: (1) has state-of-the-art certified robustness to patch attacks in the ImageNet setting (2) has high standard accuracy comparable to a standard (non-smoothed) ResNet-50, and (3) is only 5x slower than it. 

%% file: sections/speed.tex
% Smoothed classifiers traditionally require extraordinarily high computation at inference time to certify robustness---up to 100k evaluations of the base classifier in the original randomized smoothing algorithm \citep{cohen2019certified}. 
% To dramatically reduce this cost, derandomized smoothing uses deterministic, structured ablation sets such as column ablations instead of randomly sampling ones \citep{levine2020randomized}. 

Derandomized smoothing with column ablations is an expensive operation, especially for large images. Indeed, an image with $h \times w$ pixels has $w$ column ablations, so the forward pass of smoothed model is $w$ times slower than a normal forward pass---\textit{two orders of magnitude} slower on ImageNet. 
%One challenge of derandomized smoothing is its high computational cost---a forward pass through the smoothed model requires passing all possible ablations through the base classifier. For column smoothing on a $h \times w$ pixel image, this is $w$ times slower than a normal forward pass. For ImageNet, this slows down evaluation by two orders of magnitude. Most of the extra computational cost is spent processing pixels in the image ablation that are blacked out.

To address this, we first modify the ViT architecture to avoid unnecessary computation on masked pixels (Section~\ref{sec:speed}). We then demonstrate that reducing the number of ablations via striding offers further speed up (Section~\ref{sec:ablation_stride}). 
These two (complementary) modifications vastly improve the inference time for smoothed ViTs, making them comparable in speed to standard (non-robust) convolutional architectures. 
% In this section, we propose a variant of the ViT which significantly speeds up evaluation of image ablations by avoiding unnecessary computation on blacked out pixels. Specifically, we leverage the token-based view of vision transformers to drop any tokens that correspond to fully masked regions of the input. We then show that this optimization substantially reduces the computational complexity of de-randomized smoothing and, in practice, can provide up to an order of magnitude speedup over CNNs.

% \subsection{Dropping masked tokens for vision transformers}
\subsection{Dropping masked tokens}
\label{sec:speed}
% \eric{added an introductory section here as well, likely needs to be revised}
% A downside of derandomized smoothing is its computational cost---a forward pass through the smoothed model requires passing all possible ablations through the base classifier. 
% For column smoothing (the fastest type of ablation for derandomized smoothing) on a $s \times s$ pixel image, this is $s$ times slower than a normal forward pass. 
% For ImageNet, this slows down evaluation by two orders of magnitude.
% \eric{Fix notation}

% One natural strategy for reducing unnecessary computation is to crop out the masked regions before passing the image to the base classifier. Although reducing the size of the image can speed up inference, cropping has several downsides for CNNs. Cropping an image strips spatial information about the ablation and disallows ablations that wrap around the sides of the image. These lead to substantial reductions in certified performance which we discuss in more detail in Appendix \ref{app:crop}. 

Recall that the first operation in a ViT is to split and encode the input image as a set of \textit{tokens}, where each token corresponds to a patch in the image. However, for image ablations, a large number of these tokens correspond to fully masked regions of the image. 

Our strategy is to pass only the \textit{subset} of tokens that contain an unmasked part of the original image, thus avoiding computation on fully masked tokens. 
Specifically, given an image ablation, we alter the ViT architecture to do the following steps: 
% Vision transformers operate on \textit{tokens}, where each token corresponds to a positionally encoded region in the image. 
% Thus, unlike CNNs, ViTs do not require their inputs to be spatially contiguous, instead process images as a set of tokens \saachi{bag of tokens?}. 
% After encoding the tokens, we can therefore pass the self-attention layers only the \textit{subset} of tokens which contain a visible part of the original image, avoiding computation on the masked regions.  Specifically, we adjust the ViT to perform the following procedure given an image ablation:
\begin{enumerate}
    \item Positionally encode the entire ablated image into a set of tokens.
    \item Drop any tokens that correspond to a \textit{fully} masked region of the input.
    \item Pass the remaining tokens through the self-attention layers.
\end{enumerate}
As one would expect, since the positional encoding maintains the spatial information of the remaining tokens, the ViT's accuracy on image ablations barely changes when we drop the fully masked tokens. We defer a detailed analysis of this phenomenon, along with a formal description of the token-dropping procedure to Appendix~\ref{app:droptokens}.

% Note that, due to the positional encoding, dropping masked tokens does not lose any spatial information. 

% Since the tokens are positionally encoded with their spatial location in the original image, we do not lose any information from the input by performing this optimization. Indeed, we confirm that dropping masked tokens does not hurt performance (Figure \ref{fig:missingness-ablation-acc}), and amounts to a free speed-up for ViTs. 

% \subsection{Computational complexity}
\paragraph{Computational complexity.}
We now provide an informal summary of the computational complexity of this procedure, and defer a formal asymptotic analysis to Appendix \ref{app:complexity}. 
% \eric{what is the hidden dimension here? I don't think we mentioned this before, we might need to include a sentence about this in the background.}
After tokenization, the bulk of a ViT consists of two main operation types:
\begin{itemize}
    \item \textit{Attention operators}, which have costs that scale quadratically with the number of tokens but linearly in the hidden dimension.
    \item \textit{Fully-connected operators}, which have costs that scale linearly with the number of tokens but quadratically in the hidden dimension. 
\end{itemize}    
Reducing the number of tokens thus directly reduces the cost of attention and fully connected operators at a quadratic and linear rate, respectively. 
For a small number of tokens, the linear scaling from the fully-connected operators tends to dominate. 
The cost of processing column ablations thus scales linearly with the width of the column, which we empirically validate in Figure \ref{fig:speed}. Further details about how we time these models can be found in Appendix \ref{app:compute}.
%Table \ref{tab:speed}. 

\begin{figure}[!htbp]
    \centering
    \includegraphics[width=.6\textwidth]{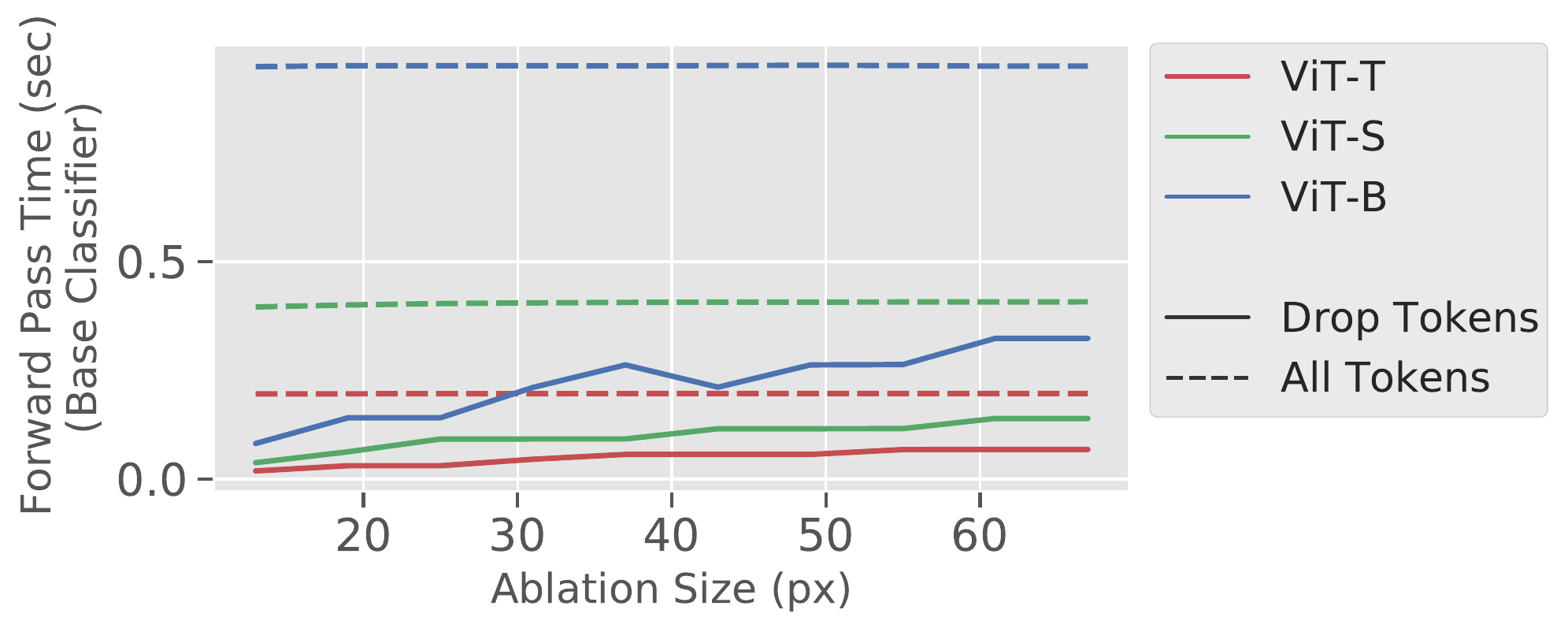}
    % \begin{subfigure}[b]{.8\textwidth}
    % \centering
    %     \includegraphics[width=.48\textwidth]{figures/Finetuned/cifar10_cert_acc.pdf}
    %     \includegraphics[width=.48\textwidth]{figures/Finetuned/cifar10_smooth_acc.pdf}
    %     \caption{CIFAR-10 models.}
    % \end{subfigure}
    % \begin{subfigure}[b]{.8\textwidth}
    %     \centering
    %         \includegraphics[width=.48\textwidth]{figures/Finetuned/imagenet_cert_acc.pdf}
    %         \includegraphics[width=.48\textwidth]{figures/Finetuned/imagenet_smooth_acc.pdf}
    %         \caption{ImageNet models.}
    %     \end{subfigure}
    \caption{
    The average time to compute a forward pass for ViTs on $1024$ column ablated images with varying ablation sizes, with and without dropping masked tokens. 
    The cost of processing a full image without dropping masked tokens corresponds to the maximum ablation size $b=224$.
    % \textbf{Test-time ablation size analysis:} Certified and clean accuracies of various \textbf{finetuned} DeiTs and comparable size ResNet models on CIFAR-10 and ImageNet. DeiTs outperforms ResNets by a significant margin especially no ImageNet while maintaining competitive clean accuracies with standard trained SOTA models. \ericinline{we need to fix this caption, it has nothing to do with the section.}
    }
    \label{fig:speed}
\end{figure}

\subsection{Empirical speed-up for smoothed ViTs}
Smoothed classifiers must process a large number of image ablations in order to make predictions and certify robustness. Consequently, using our ViT (with dropped tokens) as the base classifier for derandomized smoothing directly speeds up inference time. In
this section, we explore how much faster smoothed ViTs are in practice. 

% % We find that smoothed ViTs have significantly improved inference times over smoothed ResNets. 
% We first measure the number of ablations per second that ViTs and ResNets can process. 
% We focus on three sizes of vision transformers---ViT-Tiny (Vit-T), ViT-Small (Vit-S), and ViT-Base (Vit-B) models \citep{rw2019timm, dosovitskiy2020image}.
% We compare to residual networks of \textit{similar size}---ResNet-18, ResNet-50 \citep{he2016deep}, and Wide ResNet-101-2 \citep{zagoruyko2016wide}, respectively. All timing experiments were done on an NVIDIA A100 with floating point precision. 

% By only processing visible parts of the image, smoothed ViTs can significantly improve over the inference time of smoothed ResNets. In Figure \ref{fig:scalability} we measure the speed of processing a batch of image ablations in both training (forward and backward) and evaluation (only forward) settings across multiple ablation sizes. We find that smoothed ViTs are vastly faster than similarly sized ResNets. 

\begin{table}
    \begin{center}
    \caption{Multiplicative speed up of inference for a smoothed ViT with dropped tokens over a smoothed ResNet, measured over a batch of 1024 images with $b=19$.}
    % \caption{We measure the average speed (over 50 trials) of computing a forward pass on 1024 column ablated images with ablation size $b=19$. We report the multiplicative speed up of using a vision transformer (dropping masked tokens) over a ResNet).}
    \label{tab:speed_resnet}
    \begin{tabular}{c|ccc}
    \toprule
    & ResNet-18 & ResNet-50 & WRN-101\\
    \midrule
    ViT-T & \textbf{5.85x} & 21.96x & 101.99x \\
    ViT-S & 2.85x & \textbf{10.68x} & 49.62x \\
    ViT-B & 1.26x & 4.75x & \textbf{22.04x} \\
    \bottomrule
    \end{tabular}
    \end{center}
\end{table}

We first measure the number of images per second that smoothed ViTs and smoothed ResNets can process. We use column ablations of size $b=19$ on ImageNet, following \citet{levine2020randomized}. 
In Table \ref{tab:speed_resnet} that describes our results, we find speedups of 5-22x for smoothed ViTs over smoothed ResNets of similar size, with larger architectures showing greater gains. 
Notably, using our largest ViT (ViT-B) as the base classifier is 1.25x faster than using a ResNet-18, despite being 8x larger in parameter count. 
Dropping masked tokens thus substantially speeds up inference time for smoothed ViTs, to the point where using a large ViT is comparable in speed to using a small ResNet. 

% We first measure the number of image ablations per second that each ViT and ResNet architecture can process, when used as a base classifier for a smoothed model. We consider column ablations of size $b=19$ on ImageNet, following \citep{levine2020randomized}.

% In Table \ref{tab:speed_resnet}, we find speedups of 5-22x for ViTs over ResNets of similar size, with larger architectures showing greater gains. 
% Notably, even our largest ViT (ViT-B) is 1.25x faster than a ResNet-18, despite being 8x larger in parameter count. 
% Dropping masked tokens thus enables a substantial increase in speed for ViTs when processing image ablations, to the point where large ViTs are comparable in speed to small ResNets. 

\paragraph{Strided ablations.}
\label{sec:ablation_stride}
% Up until this point, we have only leveraged the ViT architecture to speed up inference for a smoothed classifier. 
We now consider a complementary means of speeding up smoothed classifiers: directly reducing the size of the ablation set via \emph{strided} ablations.  
Specifically, instead of using every possible ablation, we can subsample every $s$-th ablation for a given stride $s$. 
Striding can reduce the total number of ablations (and consequently speed up inference) by a factor of $s$, \emph{without} substantially hurting standard or certified accuracy (Table~\ref{table:main_summary table}). We study this in more detail in Appendix~\ref{app:stride}. 
% In Appendix \ref{app:stride}, 

% The bulk of the runtime of derandomized smoothing comes from aggregating predictions over a large number of ablations. 
% For example, column smoothing on ImageNet aggregates the result over $224$ column ablations, slowing down inference by two orders of magnitude. 
% Although we previously sped this process up by dropping tokens for ViTs, another possibility is to use strided ablations to reduce the total number of required inference steps for certification. 
% Specifically, instead of using every possible ablation, we can instead take every $s$-th ablation for a given stride $s$. 
% Strided ablations reduce the total number of ablations (and consequently speed up inference) by a factor of $s$, without substantially changing the certification threshold (up to rounding). 
% \saachi{do we need this figure? can we just put something like: with stride 10 we get a 10x speedup with < X\% change in clean/certified acc}

% \begin{example}
% For column smoothing with width $b$ and stride $s$, the maximum number of ablations that an $m \times m$ patch can intersect with is at most $\Delta_{column+stride} = \lceil(m+s-1)/s\rceil$. 
% \end{example}

% We find that striding does not substantially change the accuracy of the ViT at classifying ablations, as shown in Figure \ref{fig:stride-sweep}. 
% For example, ViT-B is $46.7\%$ accurate at classifying ablations with stride $s=10$---an order of magnitude faster than using all column ablations for only a $0.3\%$ drop in ablation accuracy. 
Strided ablations, in conjunction with the dropped tokens optimization from Section \ref{sec:speed}, lead to smoothed ViTs having inference times comparable to standard (non-robust) models. For example, when using stride $s=10$ and dropping masked tokens, a smoothed ViT-S is only 2x slower than a single inference step of a standard ResNet-50, while a smoothed ViT-B is only 5x slower.  We report the inference time of these models, along with their standard and certified accuracies, in Table \ref{table:main_summary table}. 

% We test this for various ViTs on ImageNet, with column ablations of width $b=19$, against $32\times 32$ patches. 
% We find multiple strides that achieve similar performance to the stride of length 1 (e.g. stride of 5 or 10) as shown in Figure~\ref{fig:stride-sweep}. 
% These results comes at 5x and 10x speedups, respectively, with slightly lower clean accuracy as shown in the right plot of Figure~\ref{fig:stride-sweep}. 
% \eric{Hadi, can you help replace the right plot with a column smoothing equivalent? And move the current block smoothing one to the appendix?}
% We find that striding can lead to high variance in performance: certified and standard accuracy can drastically shift either upwards or downwards as we vary the stride. 
% % \eric{we should probably say something about the non-monotonic nature of the certified accuracy}
% However, as long as the stride is carefully selected, we can trade-off a slight decreases in certified and clean accuracy for significantly faster certification. 
% \eric{Saachi, how close is this to a standard forward pass of a non-smoothed model?}

%% file: sections/related.tex
\paragraph{Certified defenses.}
An extensive body of research has studied the development of certified or provable defenses to adversarial perturbations. 
This line of research largely falls into one of three categories: tighter or exact verifiers \citep{katz2017reluplex, ehlers2017formal, lomuscio2017approach, tjeng2019evaluating, xiao2019training}, convex relaxation-based defenses \citep{wong2018provable, raghunathan2018certified, wong2018scaling, gowal2018on, gowal2019scalable, mirman2018differentiable, weng2018towards, zhang2018efficient, salman2019convex}, and smoothing-based defenses \citep{lecuyer2018certified, li2018certified, cohen2019certified,salman2019provably,levine2020wasserstein, levine2020randomized, yang2020randomized, salman2020denoised}. 
In the case of patches, the earliest certified defense used an instance of convex relaxation (interval bounds) to derive provable guarantees to adversarial patch \citep{chiang2020certified}. 
Subsequent work \citep{levine2020robustness} focused on randomized smoothing. This approach smooths classifiers over random noise, but tend to be extremely expensive to use (4-5 orders of magnitudes slower than a standard, non-robust model) \citep{cohen2019certified, levine2020robustness}. Recently, \citet{lin2021certified} proposed a variant based on randomized cropping that performs similarly to \citet{levine2020randomized} but with better guarantees under worse-case patch transformations. 

\paragraph{Deterministic smoothing.} To mitigate the expensive inference times of randomized smoothing, \citet{levine2020randomized} proposed derandomized smoothing, which used a finite set of ablations to smooth a base classifier. This substantially reduced the computational requirements of smoothing, but is still two orders of magnitude slower than standard models. Two similar defenses, Clipped BagNet \citep{zhang2020clipped} and PatchGuard \citep{xiang2021patchguard}, rely on restricting the model's receptive field. These approaches are faster than derandomized smoothing, but have other limitations. 
Clipped BagNet has substantially weaker robustness guarantees than derandomized smoothing. PatchGuard has higher but \textit{brittle} guarantees: a defended model is optimally defended against a specific patch size, and achieves no robustness at all against patches that are even slightly larger than the one considered. 

% \paragraph{Certified} Below are recent certified patch defenses.

%The first robustness guarantee against adversarial patches was proposed by \cite{} which utilized a technique, previously used to provably defend against $\ell_p$-attacks, called Interval Bound Propagation (IBP) \cite{}. 
%Afterwards, Clipped BagNet (CBN) \cite{zhang2020clipped}, De-randomized Smoothing (DS) \cite{levine2020randomized}, Minority Report (MR) \cite{mccoyd2020minority}, PatchGuard \cite{xiang2021patchguard}, and BagCert~\cite{metzen2021efficient} were proposed to improve provable security guarantee against the adversarial patch attack. PatchGuard++ \cite{xiang2021patchguard++} is the latest defense that outperforms previous provably robust defenses. It can be viewed as a hybrid of PatchGuard (small receptive fields + feature masking) and MR (checking consensus of masked predictions).

%\subsection{Certified robustness}

%\paragraph{Convex-relaxatin based methods}
%\paragraph{Randomized smoothing}

%\subsection{Patch attacks and defenses}

%\subsubsection{Attacks.}

\paragraph{Empirical methods: attacks and defenses.}
Another line of work studies empirical approaches for generating adversarial patches and designing empirical defenses. 
Adversarial patches have been developed for downstream tasks such as image 
classification \citep{karmon2018lavan}, object detection \citep{eykholt2018physical, chen2018shapeshifter, liu2018dpatch}, and facial recognition \citep{sharif2016accessorize, thys2019fooling, bose2018adversarial}. Several of these attacks work in the 
physical domain \citep{brown2018adversarial, eykholt2018physical, chen2018shapeshifter}, and 
can successfully target tasks such as traffic sign recognition \citep{eykholt2018physical, chen2018shapeshifter}. Heuristic defenses to these attacks include watermarking \citep{hayes2018visible} and gradient smoothing \citep{naseer2019local}; however, these defenses were shown to be vulnerable adaptive attacks \citep{chiang2020certified}. More recently, \citet{rao2020adversarial} proposed an adversarial training approach to improve empirical robustness to patch attacks. 

%\citet{brown2018adversarial} first proposed a universal targeted adversarial patch attack in the physical world to hijack the model prediction. 

%\cite{karmon2018lavan} later proposed a more powerful attack in the digital domain. More following up works also showed that the localized adversary can achieve a robust physical attack against traffic sign recognition, human detection \cite{eykholt2018physical,chen2018shapeshifter}.

%Because of the limitations of the classification setting, several other works have investigated the use of adversarial patches in the object detection and face recognition settings \cite{sharif2016accessorize, eykholt2018physical,thys2019fooling, chen2018shapeshifter, bose2018adversarial, liu2018dpatch}.

% \subsubsection{Defenses.}
% \paragraph{Empirical.}
% Heuristic-based Digital Watermarking \cite{hayes2018visible} 
% Local Gradient Smoothing \cite{naseer2019local}.

% However, these had been shown vulnerable to an adaptive attacker \cite{chiang2020certified}. 

% \citet{rao2020adversarial} presented adversarial training against patches with location optimization plus and gave an overview of the limited patch work.

\paragraph{Vision transformers.}
Our work leverages the vision transformer (ViT) architecture \citep{dosovitskiy2020image}, which adapts the popular attention-based model from the language setting \citep{vaswani2017attention} to the vision setting. 
Recent work \citep{touvron2020training} has released more efficient training methods as well as pre-trained ViTs that have made these architectures more accessible to the wider research community. 
% ViTs recently got a lot of attention in the research community from works ranging  efficient methods for training ViTs \citet{touvron2020training} to studying the properties of ViTs in terms of robustness for example \citep{shao2021adversarial, mahmood2021robustness}.
% [eric] I find this kind of "ours of is the first to do X and Y together" as kind of cringy 
%To our knowledge, ours is the first work to explore patch attacks and certified defenses with visual transformers. 
%Vision Transformers (ViT) were first proposed by \citet{dosovitskiy2020image} as a direct adaptation of the popular transformers architecture used in NLP applications \cite{vaswani2017attention} for computer vision. In particular, ViTs do not 
%include any convolutions, instead tokenizing the image into patches which are then passed through several full layers 
%of multi-headed self-attention. \citet{touvron2020training} further adjust the training strategies for ViT to allow data efficient transformers (DeiT) to outperform CNNs without pretraining on large datasets.  While there has been some work on exploring the adversarial robustness of visual transformers \cite{shao2021adversarial, mahmood2021robustness}, these works focus solely on $l_2$ and $l_\infty$ robustness. To our knowledge, ours is the first work to explore patch attacks and certified defenses with visual transformers. 

%% file: sections/conclusion.tex
We demonstrate how applying visual transformers (ViTs) within the smoothing framework leads to significantly improved certified robustness to adversarial patches while maintaining standard accuracies that are on par with regular (non-robust) models.
Further, we put forth changes to the ViT architecture and the corresponding smoothing procedure that greatly speed up the resulting inference times over previous smoothing approaches by up to two orders of magnitude---they end up being only 2-5x slower than that of a regular ResNet. 
We believe that these improvements finally establish models that are certifiably robust to adversarial patches as a viable alternative to standard (non-robust) models.

%% file: sections/acknowledgement.tex
Work supported in part by the NSF grants CCF-1553428 and CNS-1815221, and Open Philanthropy. This material is based upon work supported by the Defense Advanced Research Projects Agency (DARPA) under Contract No. HR001120C0015.

Research was sponsored by the United States Air Force Research Laboratory and the United States Air Force Artificial Intelligence Accelerator and was accomplished under Cooperative Agreement Number FA8750-19-2-1000. The views and conclusions contained in this document are those of the authors and should not be interpreted as representing the official policies, either expressed or implied, of the United States Air Force or the U.S. Government. The U.S. Government is authorized to reproduce and distribute reprints for Government purposes notwithstanding any copyright notation herein.

%% file: sections/app-setup.tex
\subsection{Models and architectures}
We use three sizes of vision transformers---ViT-Tiny (ViT-T), ViT-Small (ViT-S), and ViT-Base (ViT-B) models \citep{rw2019timm, dosovitskiy2020image} and compare to to residual networks of similar (or larger) size---ResNet-18, ResNet-50 \citep{he2016deep}, and Wide ResNet-101-2 \citep{zagoruyko2016wide}, respectively. These architectures and their corresponding number of parameters are summarized in Table~\ref{table:architectures}.

\begin{table}[!htbp]
    \caption{A collection of neural network architectures we use in our paper.}
    \centering
    \begin{tabular}{l|cccccccc}
        \toprule
        Architecture & ViT-T & ResNet-18 && ViT-S & ResNet-50 && ViT-B & WRN-101-2 \\
        Params & 5M & 12M && 22M & 26M && 86M & 126M \\
        \bottomrule
    \end{tabular}
    \label{table:architectures}
\end{table}

We use the same architectures for both ImageNet and CIFAR-10 models, and finetune our smoothed models from publicly released checkpoints pretrained on ImageNet. All our CIFAR-10 experiments are thus conducted on up-sampled CIFAR-10 images of size $224\times224$. 

\subsection{Datasets} 
\label{app:datasets}
We use two datasets: 
\begin{enumerate}
    \item CIFAR \citep{krizhevsky2009learning} \url{https://paperswithcode.com/dataset/cifar-10}.
    \item ImageNet \citep{russakovsky2015imagenet}, with a custom (research, non-commercial) license, as found here \url{https://paperswithcode.com/dataset/imagenet}.
\end{enumerate}

\subsection{Training parameters}
\label{app:training}
Derandomized smoothing requires that the base classifier predict well on image ablations.
A standard technique for derandomized smoothing methods is to directly train the base classifier on image ablations \citep{levine2020randomized}. Thus, unless otherwise stated, in each epoch we randomly apply a column ablation of fixed width to each image of the training set.
% Derandomized smoothing requires the base classifiers to perform well on image ablations. Therefore, a standard way in the literature to achieve is via data-augmentation, i.e., train on image ablations \cite{levine2020randomized}. We thus use this technique in our paper. Unless otherwise stated, in each epoch we randomly apply column ablation of fixed width to each image of the training set.

To facilitate training of the base classifiers, we start from pretrained ResNets\footnote{These are TorchVision's official checkpoints, and can be found here \url{https://pytorch.org/vision/stable/models.html}.} and ViT architectures\footnote{We use the DeiT checkpoints of \cite{rw2019timm} which can be found here \url{https://github.com/rwightman/pytorch-image-models/blob/master/timm/models/vision_transformer.py}.} and fine-tune as follows:

\paragraph{ImageNet.}
We train for $30$ epochs using SGD of fixed learning rate of $10^{-3}$, a batch size of $256$, a weight-decay of $10^{-4}$, a momentum of $0.9$, and with column ablations of fixed width $b=19$. For data-augmentation, we use random resized crop, random horizontal flip, and color jitter. We then apply column ablations.

\paragraph{CIFAR-10.}
We train for $30$ epochs using SGD with a step learning rate of $10^{-2}$ that drops every 10 epochs by a factor of 10, a batch size of $128$, a weight-decay of $5 \times 10^{-4}$, a momentum of $0.9$, and with column ablations of fixed width $b=4$. We only use random horizontal flip for data-augmentation, after which we apply column ablations. We then upsample all CIFAR-10 images to $224\times224$ (on GPU). 

\paragraph{Training time.}
Training is relatively fast, with our largest ImageNet model (WRN-101-2) finishing in roughly two days on one NVIDIA V100 GPU. The smaller models such as ViT-T or ResNet-18 finish training in only a few hours.

\subsection{Compute and timing experiments} 
\label{app:compute}
We use an internal cluster containing NVIDIA 1080-TI, 2080-TI, V100, and A100 GPUs. Scalability and timing experiments were performed on an A100 
and averaged over 50 trials. When performing scalability experiments, we do not include data loading time or the time to move the input to the GPU. 

\subsection{Example ablations}
\begin{figure}[!htbp]
    \centering
    \begin{subfigure}[b]{.7\textwidth}
        \centering
        \includegraphics[width=\textwidth]{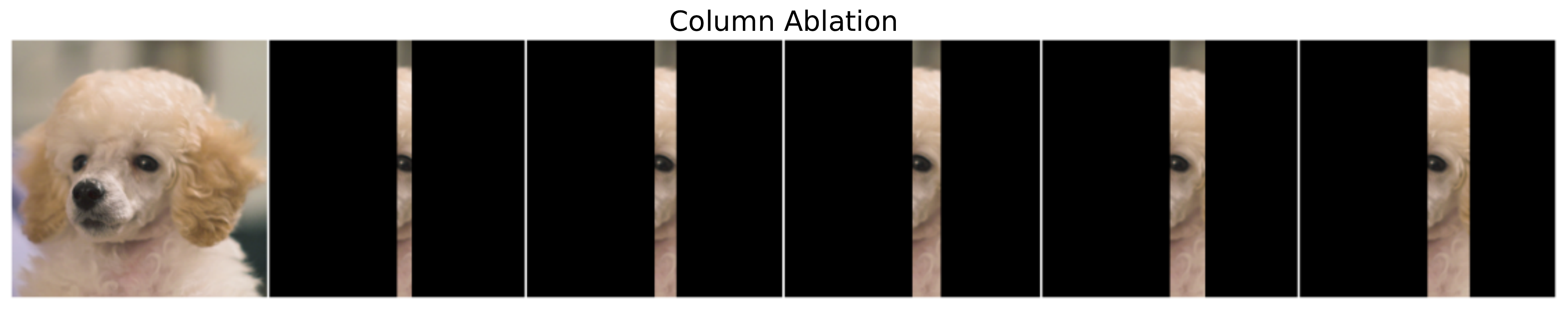}
        \caption{\textbf{Column ablations} with the following ablation size from left to right: original image, 13px, 19px, 25px, 31px, 37px.}
        \end{subfigure}
        \begin{subfigure}[b]{.7\textwidth}
            \centering
                \includegraphics[width=\textwidth]{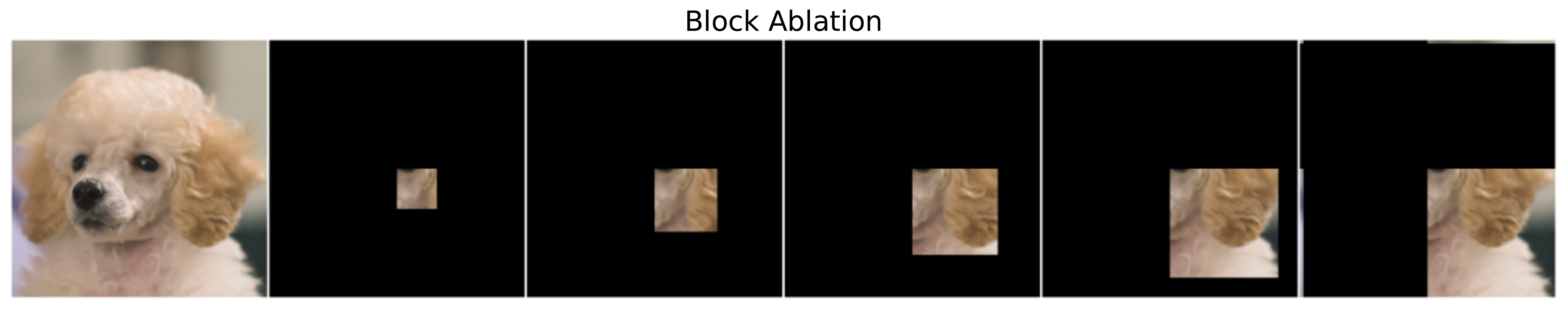}
                \caption{\textbf{Block ablations} with the following ablation size from left to right: original image, 35px, 55px, 75px, 95px, 115px.}
            \end{subfigure}
        \caption{Example ablations that we use in our paper.}
    \label{appfig:example-ablation-sizes}
\end{figure}

In Figure~\ref{appfig:example-ablation-sizes}, we display examples of ablations of various types (column, block) and sizes.

\subsection{Differences in setup from \citet{levine2020randomized}}
\label{app:differences}
Our work builds on top of that of \citet{levine2020randomized}. We use their robustness guarantee as is (see Section~\ref{sec:preliminaries}), but there are a few differences in the setup of our experiments. 
All experimental results (including the de-randomized smoothing baseline) are run using the same experimental setup in order to remain fair, which only improved the baseline over what was previously reported in the literature. For completeness, we describe the differences in setup here. 

\paragraph{Encoding \textit{null} inputs.} The first difference is that \citet{levine2020randomized} encode part of the input as being \textit{null} or ablated by adding additional color channels, as described in \cite{levine2020robustness}, so that the \textit{null} value is distinct from all real pixel colors. 
In practice, we found this to be unnecessary, and were able to replicate their results with ablations that simply replace masked pixels with 0. 

\paragraph{Early stopping.} We find that ResNets substantially benefit from early stopping when trained with ablations, and otherwise experience severe overfitting to the ablations with substantially reduced test accuracies. 
In our replication, we find that the ResNet-50 result reported by \citet{levine2020randomized} can be substantially improved with an earlier checkpoint (improving certified accuracy by nearly 10\%), and thus we use early-stopping in all of our ResNet baselines. 

\paragraph{Starting from pretrained models.}
To reduce training time, for both ImageNet and CIFAR-10 experiments, we start from pre-trained ImageNet checkpoints (see Section \ref{app:training}). This step is especially necessary for the CIFAR-10 experiments, as it is quite challenging to train a ViT from scratch on CIFAR-10 (these models tend to require a large amount of data).

\paragraph{Upsampled CIFAR-10.}
In order to use the pretrained ImageNet checkpoints when training our base classifiers for CIFAR-10, we (nearest neighbor) upsample the CIFAR-10 inputs to $224\times224$ as part of the model architecture. We verify robustness in the original $32\times32$ images. 

\paragraph{Sweeping over ablation size.} 
We note that \citet{levine2020randomized} tested various ablations sizes only on CIFAR-10. Due to our speed-ups, we were able to sweep over ablations sizes for ImageNet. 

%% file: sections/app-traintime-abl.tex
In this section, we further explore the impact of changing the ablation size on both standard and certified performance. In Section \ref{app:trainablations}, we explore the effect of modifying the ablation size at training time. 
In Section~\ref{app:testablations}, similar to the experiment on ImageNet from Section~\ref{sec:ablation_size},
we present additional results on adjusting the ablation size at test time for CIFAR10.

\subsection{Train-time ablation}
\label{app:trainablations}
\begin{figure}[!htbp]
    \centering
    \begin{subfigure}[b]{.48\textwidth}
    \centering
        \includegraphics[width=\textwidth]{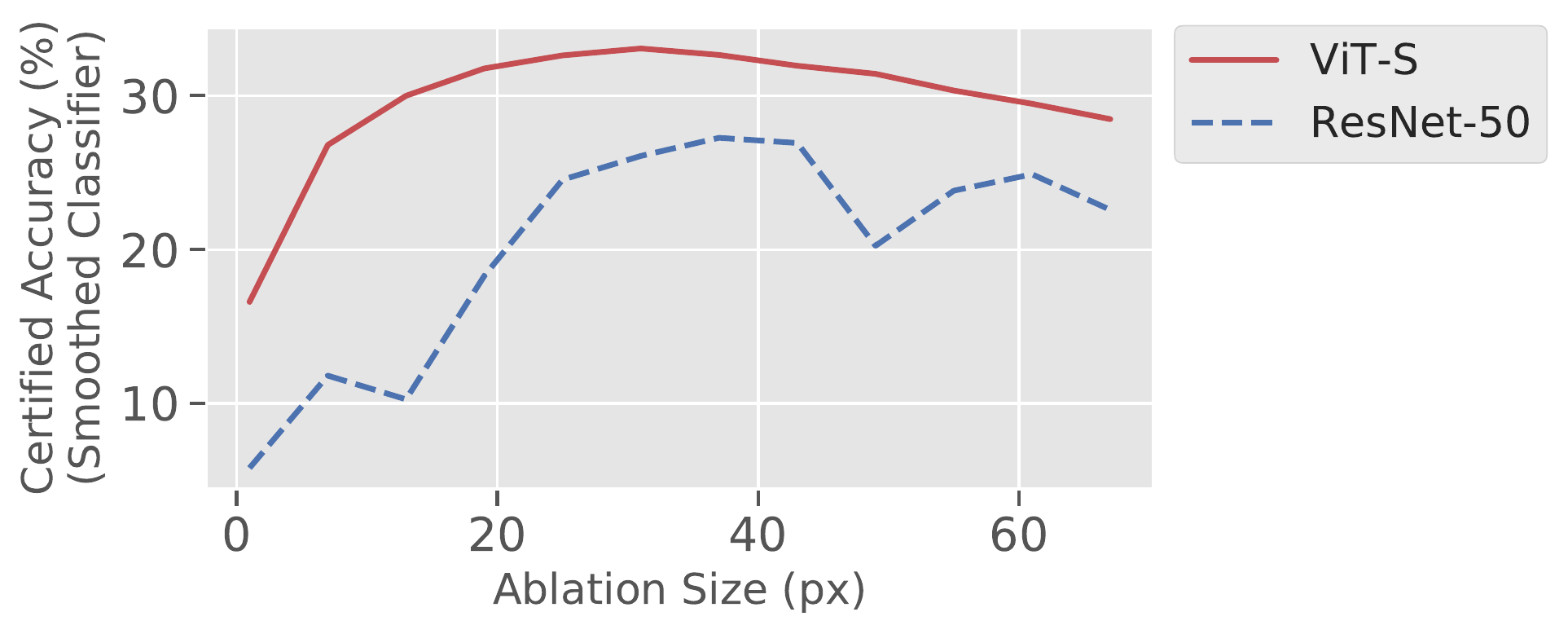}
        % \caption{Finetuned CIFAR-10 models.}
    \end{subfigure}
    \begin{subfigure}[b]{.48\textwidth}
        \centering
            \includegraphics[width=\textwidth]{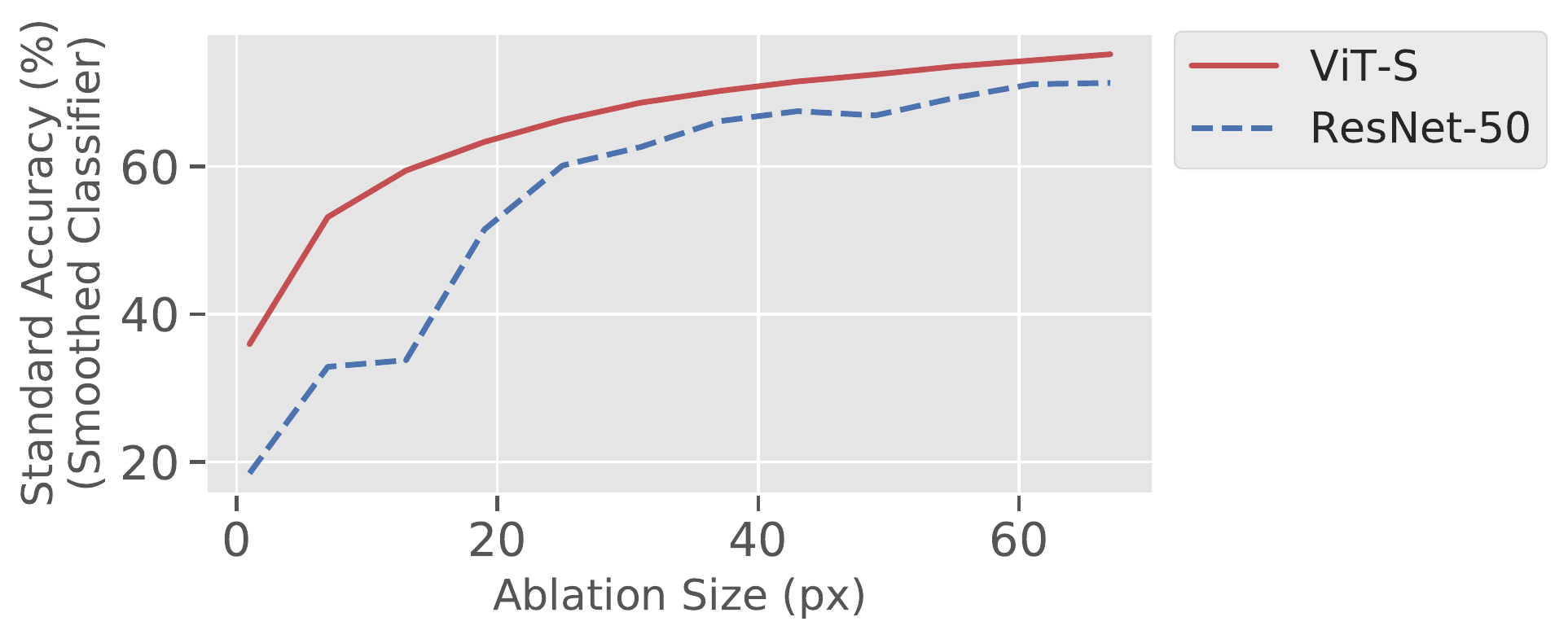}
            % \caption{Finetuned ImageNet models.}
        \end{subfigure}
    \caption{
    Certified and standard accuracy for a smoothed model trained and evaluated on ImageNet column ablations with varying widths. The ResNet-50 requires a substantially larger ablation size for certification, whereas the ViT-S is more flexible.
    }
    \label{fig:ablation-size-train-time}
\end{figure}

We first explore varying the ablation size used during training for ImageNet.
Specifically, we train and certify a ResNet-50 and ViT-S over a range of column widths from 1 to 67 pixels (Figure \ref{fig:ablation-size-train-time}).

For ViTs, we find that once the columns are wide enough, we see only marginal improvements in certified accuracy (i.e. only 1.3\% higher certified accuracy over $b=19$). 
This suggests that small ablations are sufficient at training time, allowing for fast training of ViTs when using cropped ablations.

On the other hand, ResNets require a substantially larger column width than was previously explored. Specifically, the certified accuracy of the ResNet baseline can be greatly improved from 18\% to 27\% when the ablation size is increased to $b=37$. This ablation size is optimal for the ResNet, but is still 6\% lower certified accuracy when compared to the ViT. 

Overall, we find that certified performance of ViTs on ImageNet remains largely stable with respect to the column ablation size used for training. We can thus use smaller ablation sizes during training (e.g $b=19$) to improve training speed while certifying using larger ablation sizes.

\subsection{Test-time ablations}
\label{app:testablations}

\begin{figure}[!htbp]
    \centering
    \includegraphics[width=.48\textwidth]{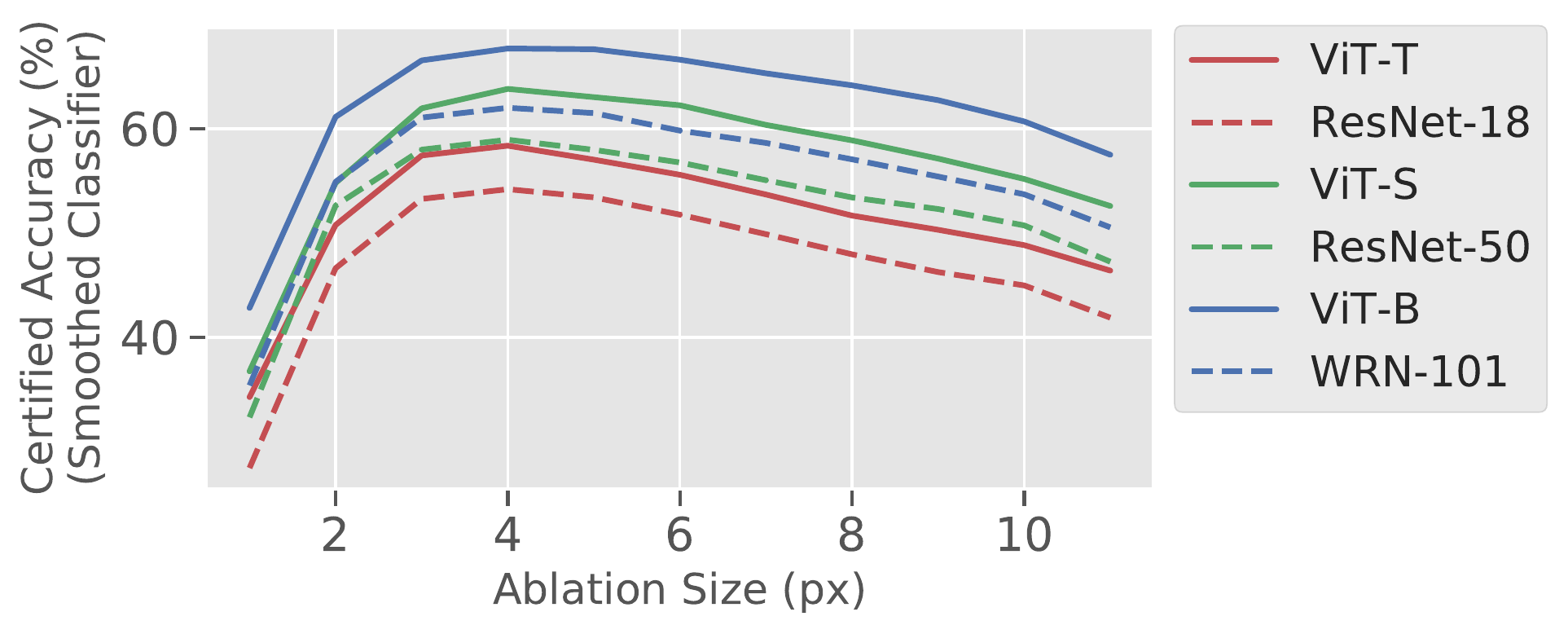}
    \includegraphics[width=.48\textwidth]{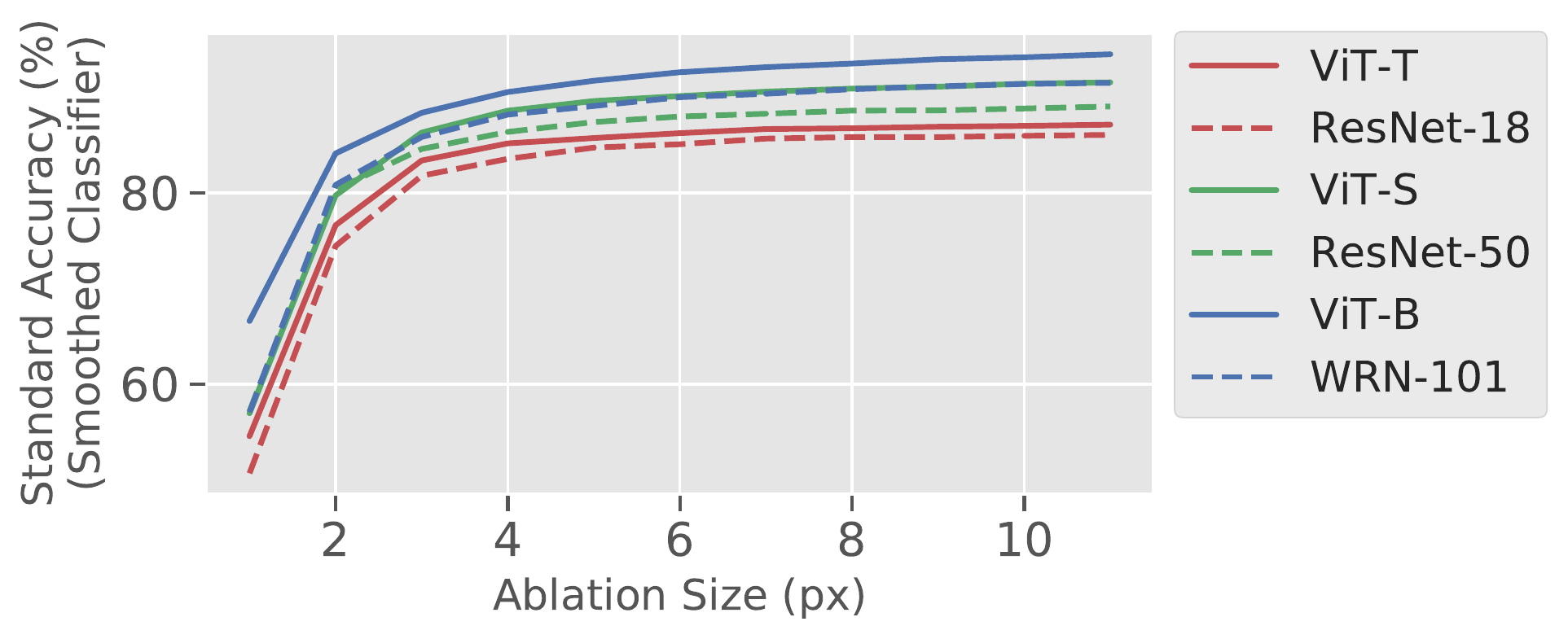}
    \caption{
    Certified and standard accuracy for a smoothed model on CIFAR-10 trained with a fixed ablation size ($b=5$), and evaluated with varying ablation sizes. 
    }
    \label{fig:ablation-size-finetuned-cifar10}
\end{figure}
Similar to the experiment on ImageNet from Section~\ref{sec:ablation_size}, we present analogous results 
for varying the ablation size used at test time for CIFAR-10. These results largely reflect what was previously observed by \citet{levine2020randomized}. Specifically, the optimal ablation size for CIFAR10 is a column width of $b=4$, with a steep drop-off in performance for larger ablation sizes. 
This is in contrast to what we observed in ImageNet, which did not see such a steep drop in performance. 

%% file: sections/app-drop-tokens.tex
We first describe the algorithm for processing image ablations with a ViT while dropping masked tokens. Let $\xin$ be an image with size $h \times w$, and let $\mathcal{S}(\mathbf{x})$ be the set of image ablations of $\xin$. For each $\mathbf{z}, \mathbf{m} \in \mathcal{S}(\mathbf{x})$, $\mathbf{z}$ is an image ablation of size $h \times w$ and $\mathbf{m}\in \{0,1\}^{h \times w}$ is its corresponding mask, such that $\textbf{m}_{ij}$ is $0$ if the $i,j$ pixel in $\mathbf{z}$ is masked and 1 otherwise.

Recall that a ViT has two stages when processing an input $\mathbf{z}$. 
\begin{itemize}
    \item \textbf{Encoding:} $\mathbf{z}$ is split into patches of $p \times p$ and positionally encoded into tokens. We let $E(\textbf{w}, i, j)$ be an encoder which positionally encodes the $p \times p$ sized patch $\textbf{w}$ which was at spatial location $ip,jp$ in $\mathbf{z}$.
    \item \textbf{Self-Attention:} A set of positionally encoded tokens $\mathcal{T}$ is passed through self attention layers $V$ and produces a class label.
\end{itemize}

Given an image ablation $\mathbf{z}$ we modify the ViT to remove tokens in $\mathcal{T}$ that correspond to a fully masked region in $\mathbf{z}$.

\begin{algorithm}
    \caption{Forward pass for processing an image ablation $\mathbf{z}$ with mask $\mathbf{m}$ using a ViT while dropping masked tokens.}
    \label{alg:drop}
    \begin{algorithmic}[1]
    \Function{ProcessAblation}{$\mathbf{z}, \mathbf{m}$}
        \State $\mathcal T = \{\}$ \textit{Initialize set of tokens for an ablation}
        \For{$i,j \in  [h/p] \times [w/p]$}
            \If{\textbf{not} $\mathbf{m}_{ip:(i+1)p, jp:(j+1)p} = \mathbf{0}$}
            \State $\mathcal{T} = \mathcal{T} \cup E(\mathbf{z}_{ip:(i+1)p, jp:(j+1)p}, i, j)$
            \EndIf
        \EndFor
        \State
        \Return $V(\mathcal{T})$
    \EndFunction
    \end{algorithmic}
\end{algorithm}

We can then use this function to define the smoothed ViT.

\begin{algorithm}
    \caption{Forward pass for a smoothed ViT on an input image $\xin$ with ablation set $\mathcal{S}(\xin)$}
    \label{alg:drop}
    \begin{algorithmic}[1]
    % \scriptsize
    \Function{SmoothedViT}{$\xin$}
    \State $c_i = 0$ for $i \in [k]$ \textit{// Initialize counts to zero}
    \For {$\mathbf{z}, \mathbf{m} \in \mathcal S(\xin)$}
        \State y = \Call{ProcessAblation}{$\mathbf{z}, \mathbf{m}$}
        \State $c_y = c_y + 1$ \textit{// Update counts}
    \EndFor
    \State 
    \Return $\argmax_{y} c_y$
    \EndFunction
    \end{algorithmic}
\end{algorithm}

\subsection{Computational complexity of ViTs with dropped tokens}
\label{app:complexity}
We can now derive the computational complexity of the smoothed ViT when dropping tokens. Specifically, consider a ViT that divides an $h\times w$ pixel image into $p \times p$ patches, and positionally encodes them tokens with $d$ hidden dimensions.

Recall that a ViT has two operation types: \textit{attention operators} which scale quadratically with the number of tokens but linearly with hidden dimension $d$ and \textit{fully-connected operators} which scale linearly with the number of tokens but quadratically in $d$. Without dropping tokens, we have $hw/p^2$ tokens. A forward pass of processing an image ablation without dropping tokens thus has an overall complexity of 
$$O\left(\left(\frac{hw}{p^2}\right)^2d + \left(\frac{hw}{p^2}\right)d^2\right)$$
where the first term corresponds to the attention operations, and the second term corresponds to the fully-connected operations. 

For column ablations with width $b$, dropping masked tokens reduces the number of tokens to $hb/p^2$. The complexity of the forward pass to process an image ablation when dropping masked tokens (i.e \texttt{ProcessAblation}) then drops to 
$$    O\left(\left(\frac{hb}{p^2}\right)^2d + \left(\frac{hb}{p^2}\right)d^2\right)$$
thus reducing the attention cost by a factor of $O(w^2/b^2)$ and the fully-connected cost by a factor of $O(w/b)$. In practice, the computation of fully-connected operations tends to dominate since $d > \frac{hw}{p^2}$. 

Overall, a smoothed ViT with stride $s$ processes $w/s$ ablations. Thus, the overall complexity of the smoothed ViT is:
$$    O\left(\frac{w}{s}\left(\left(\frac{hb}{p^2}\right)^2d + \left(\frac{hb}{p^2}\right)d^2\right)\right)$$

\subsection{Effect of dropping tokens on speed}
\begin{figure}[!htbp]
    \centering
    \begin{subfigure}[b]{0.49\textwidth}
    \centering
        \includegraphics[width=\textwidth]{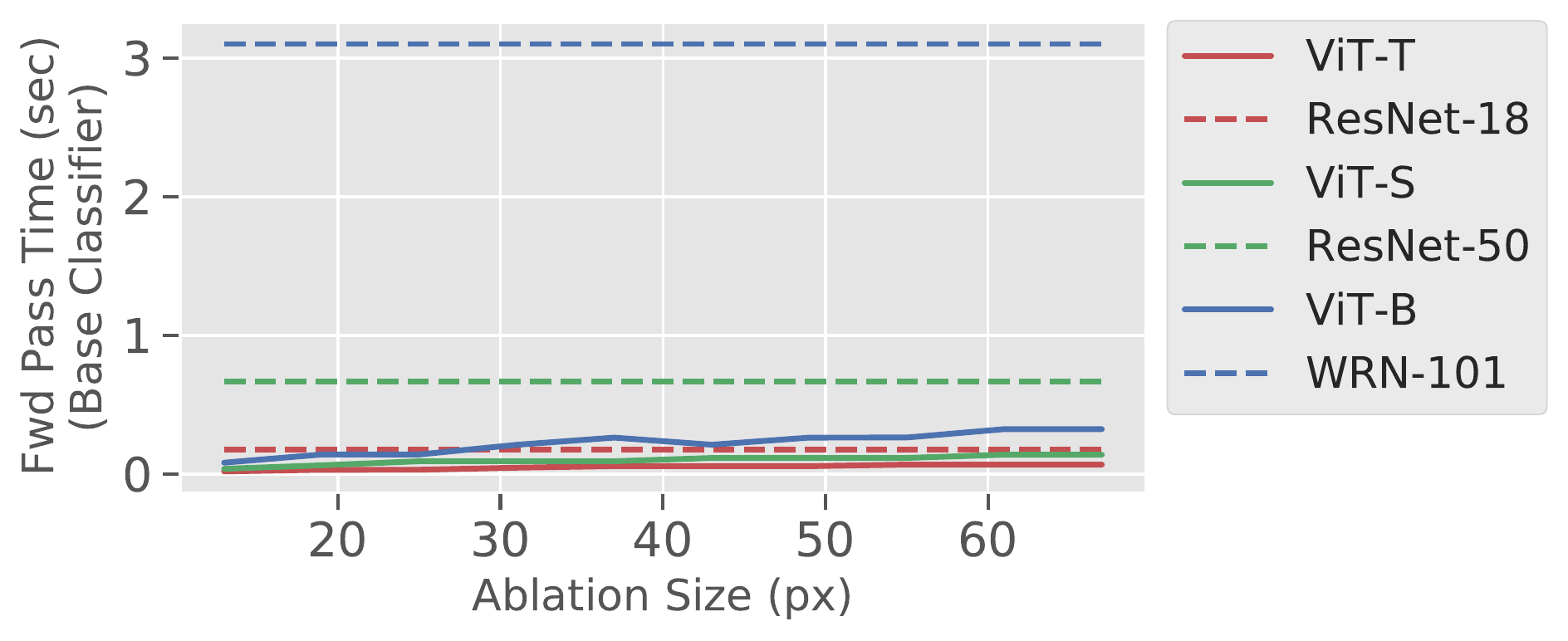}
        \caption{Forward Pass Time}
    \end{subfigure}
    \begin{subfigure}[b]{0.49\textwidth}
        \centering
        \includegraphics[width=\textwidth]{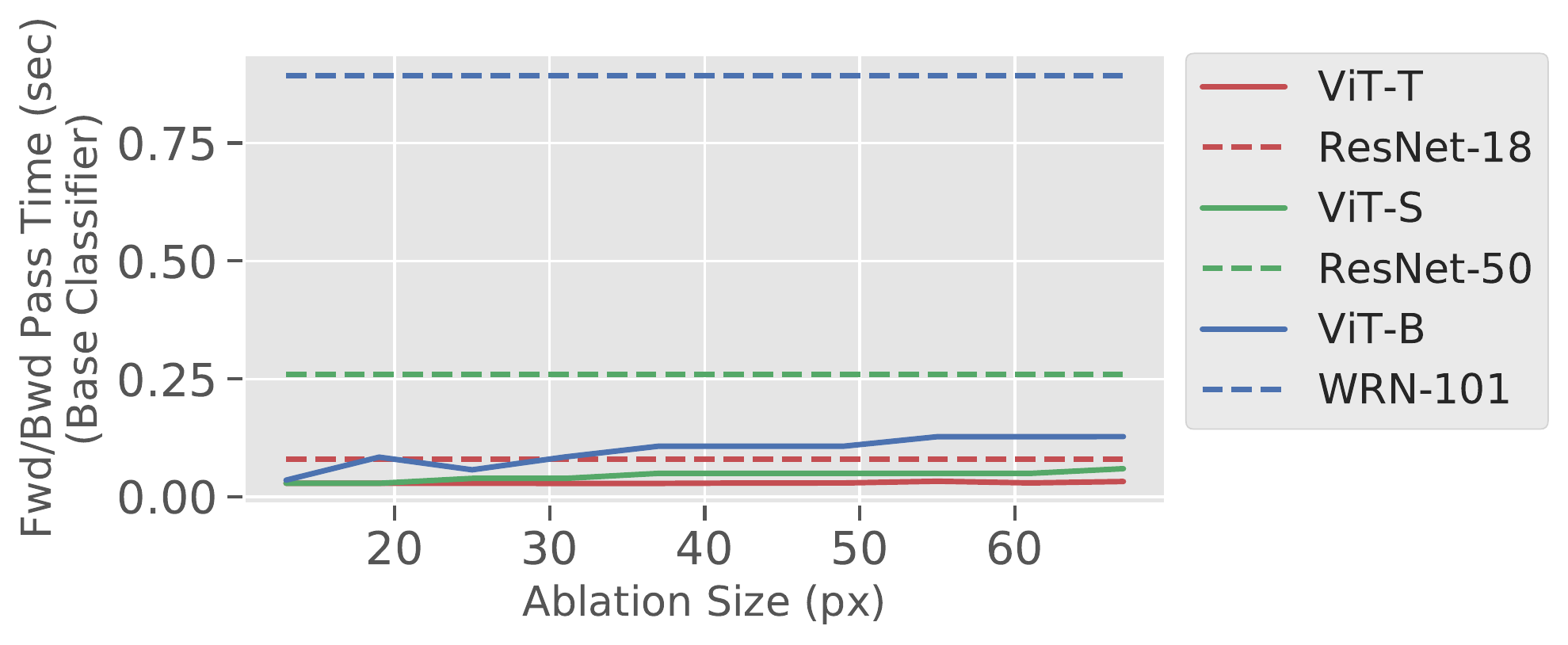}
        \caption{Forward/Backward Pass Time}
    \end{subfigure}
\caption{(a) Average time for computing a forward pass on a batch of 1024 image ablations on ImageNet (b) Average time for computing a full training step (forward and backward pass) on a batch of 128 image ablations on ImageNet}
\label{appfig:scalability}
\end{figure}

We extend the timing experiments comparing ViTs and ResNets to a range of ablation sizes (previously presented in Table~\ref{tab:speed_resnet} from Section~\ref{sec:improve-speed} for a single column ablation size of $b=19$). 
Empirically, even for substantially larger ablations, we find significantly faster training and inference times for ViTs over ResNets. In Figure \ref{appfig:scalability}, we compare the evaluation and training speeds for processing image ablations with ResNets and ViTs with dropped tokens. 

\subsection{Effect of dropping tokens on performance}
\begin{figure}[!htbp]
    \centering
    \begin{subfigure}[b]{0.49\textwidth}
    \centering
        \includegraphics[width=\textwidth]{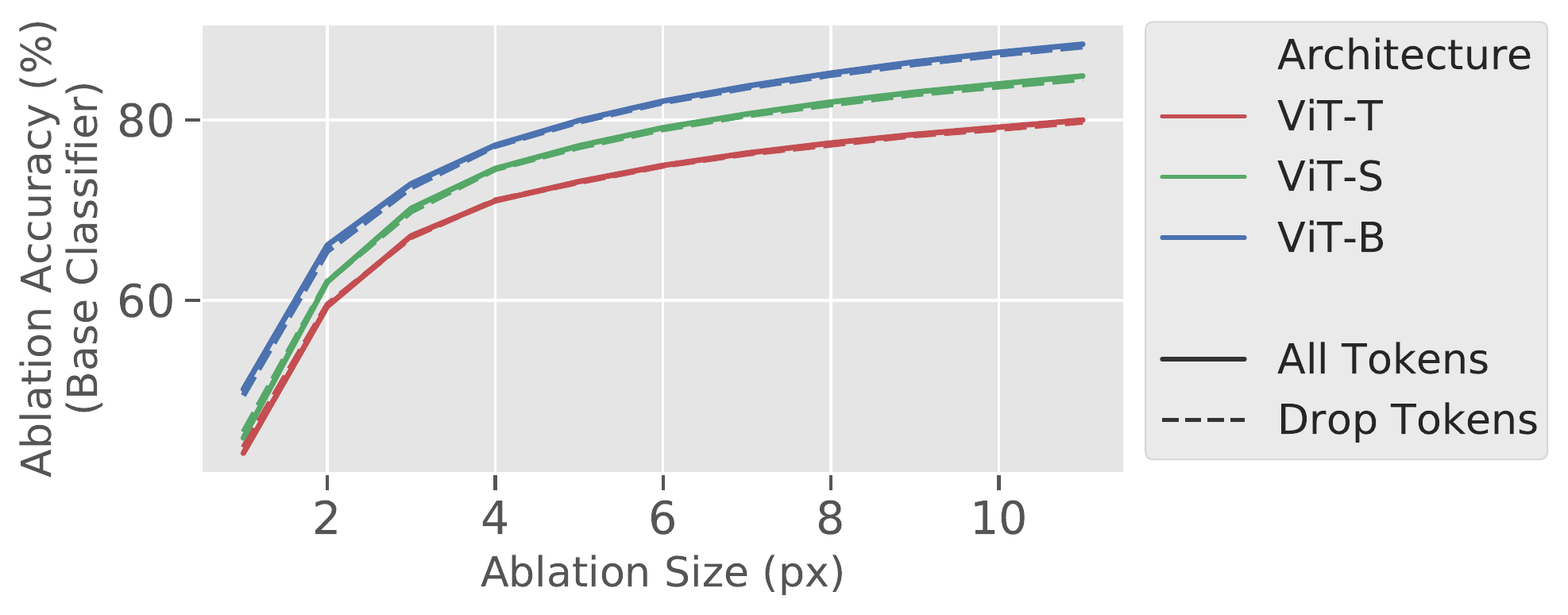}
        \caption{CIFAR-10 models.}
    \end{subfigure}
    \begin{subfigure}[b]{0.49\textwidth}
        \centering
        \includegraphics[width=\textwidth]{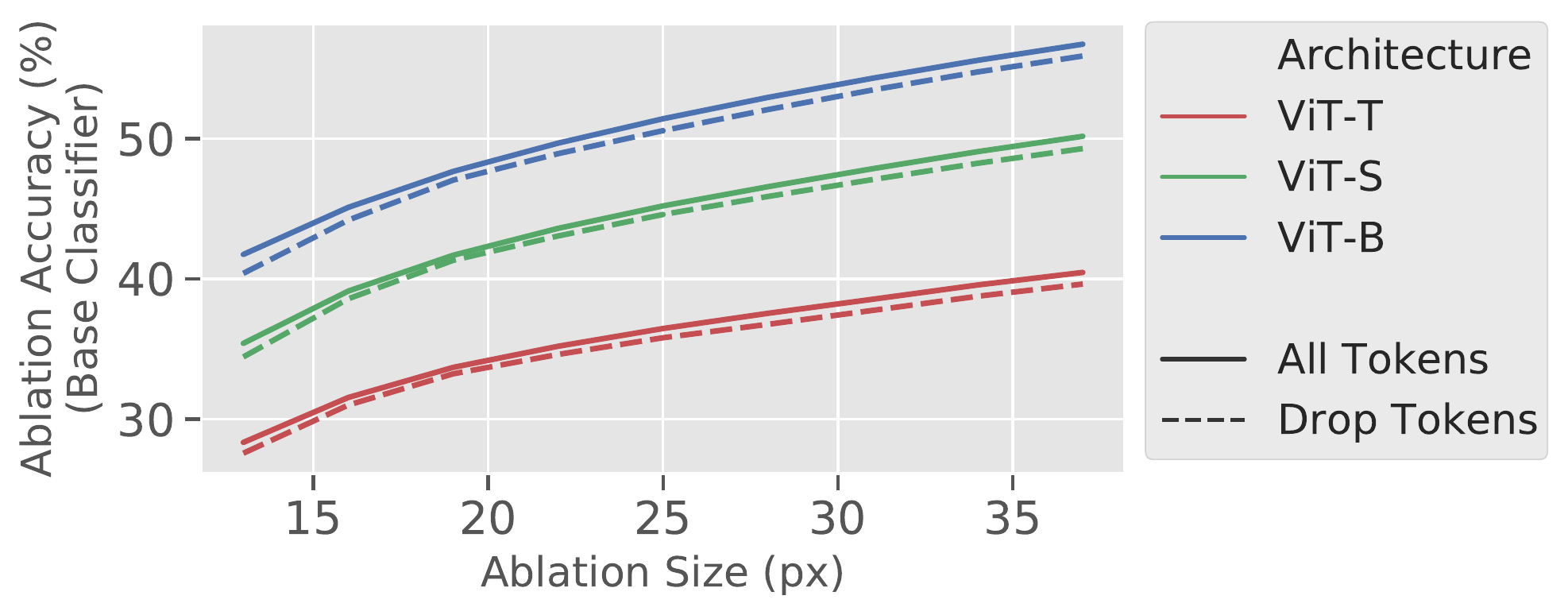}
        \caption{ImageNet models.}
    \end{subfigure}
\caption{We compare the ablation accuracies of dropping masked tokens versus processing all tokens for a collection of vision transformers on CIFAR-10 and ImageNet. Dropping masked tokens does not substantially degrade accuracy.}
\label{appfig:missingness-ablation-acc}
\end{figure}

Since the tokens are individually positionally encoded, dropping tokens that are fully masked does not remove any information from the input. In Figure \ref{appfig:missingness-ablation-acc}, we confirm that dropping masked tokens does not significantly change the accuracy of the ViT base classifier on ablations. 

% \begin{figure}[!htbp]
%     \centering
%     \begin{subfigure}[b]{.8\textwidth}
%     \centering
%         \includegraphics[width=.48\textwidth]{figures/finetuned_missingness/cifar10_cert_acc.pdf}
%         \includegraphics[width=.48\textwidth]{figures/finetuned_missingness/cifar10_smooth_acc.pdf}
%         \caption{CIFAR-10.}
%     \end{subfigure}
%     \begin{subfigure}[b]{.8\textwidth}
%         \centering
%             \includegraphics[width=.48\textwidth]{figures/finetuned_missingness/imagenet_cert_acc.pdf}
%             \includegraphics[width=.48\textwidth]{figures/finetuned_missingness/imagenet_smooth_acc.pdf}
%             \caption{ImageNet.}
%         \end{subfigure}
%     \caption{Certified and clean accuracies of various ViTs on CIFAR-10 and ImageNet, with and without missingness. The performance of both are comparable to each other while the missigness ones are 2 orders of magnitude faster.}
%     \label{appfig:missingness}
% \end{figure}

%% file: sections/app-stride.tex
In this section, we explore strided ablations for certification in more depth. In Section~\ref{app:stride_threshold} we present the threshold for certification when using strided ablations. In Section \ref{app:stride_performance} we show how striding affects performance. 

\subsection{Certification thresholds for strided ablation sets}
\label{app:stride_threshold}
We briefly describe the new thresholds for certification when using strided ablations. Recall from \eqref{eq:certify} that a prediction is certified robust if 
$$n_c(\mathbf{x}) > \max_{c' \neq c} n_{c'}(\mathbf{x}) + 2\Delta.$$
Thus $\Delta$, the number of ablations that a patch can intersect, fully describes the certification threshold. 

\paragraph{Column smoothing.} For column smoothing with width $b$ and stride $s$, the maximum number of ablations that an $m \times m$ patch can intersect with is at most $\Delta_{column+stride} = \lceil(m+s-1)/s\rceil$. 

% \paragraph{Block smoothing.} For block smoothing with block length $b$ and stride $s$, the maximum number of ablations that an $m \times m$ patch can intersect with is at most $\Delta_{column+stride} = \lceil(m+s-1)/s\rceil^2$. 

\subsection{Performance under strided ablations}
\label{app:stride_performance}
\begin{figure}[!htbp]
    \centering
    \begin{subfigure}[b]{.49\textwidth}
    \centering
        \includegraphics[width=\textwidth]{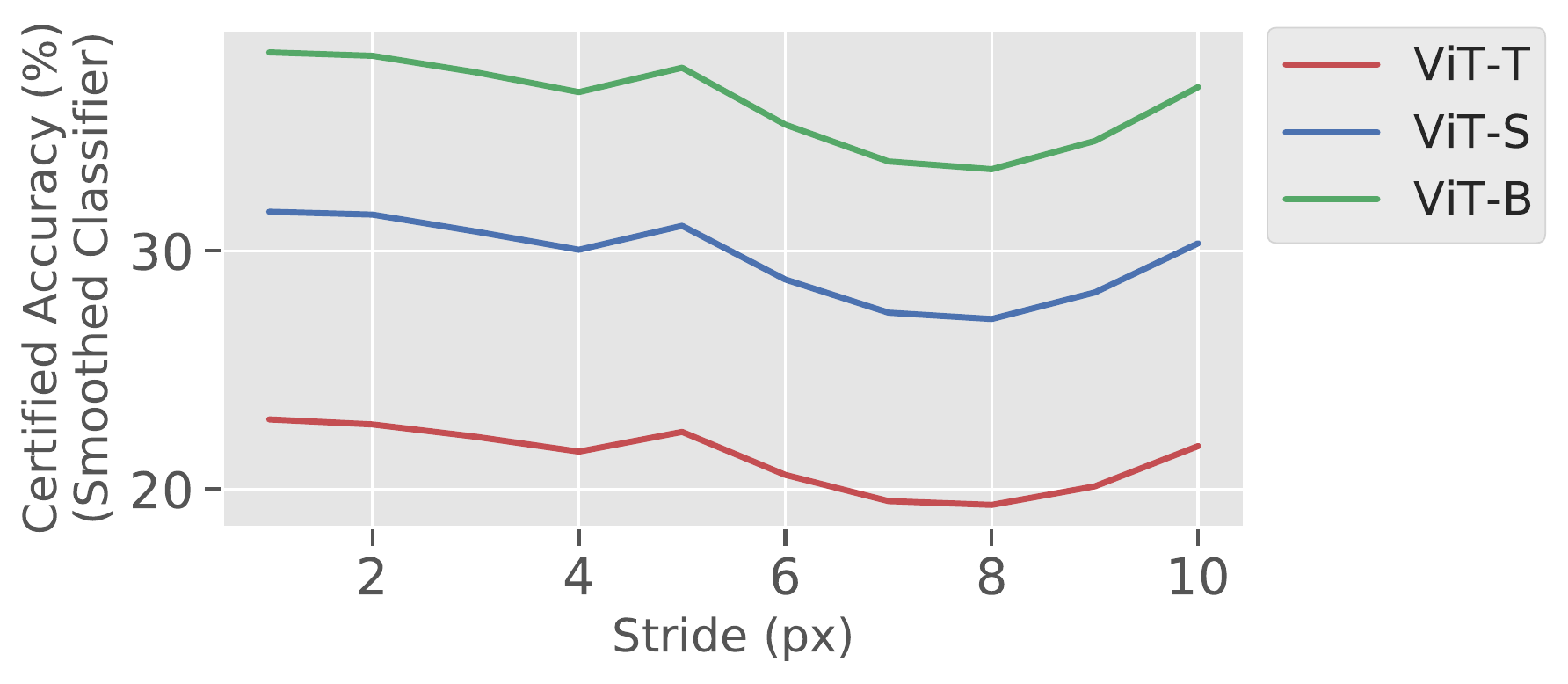}
        % \caption{Pretrained CIFAR-10 models.}
        \caption{Certified Accuracy}
    \end{subfigure}
    \begin{subfigure}[b]{.49\textwidth}
        \centering
            \includegraphics[width=\textwidth]{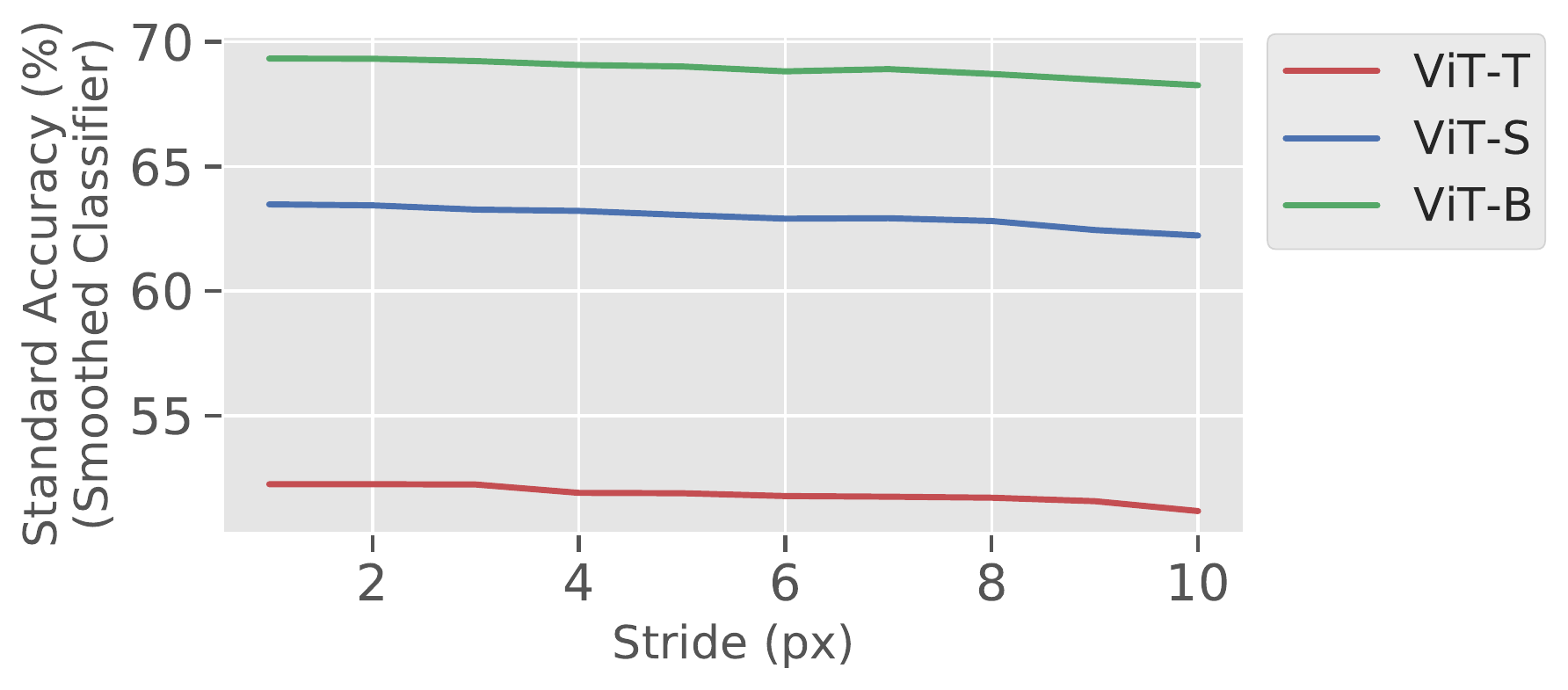}
            % \caption{Pretrained ImageNet models.}
            \caption{Standard Accuracy}
        \end{subfigure}
    \caption{Certified and standard accuracy of various ViTs for ImageNet when using strided column ablations with varying stride lengths.}
    \label{appfig:stride-sweep}
\end{figure}
% \eric{Hadi, can you help redo this figure to be just ablation accuracy? And maybe place side by side with the speed-up table to save on vertical space?}
In this section, we explore how striding affects standard and certified performance. We find that striding does not result in a monotonic change in certified accuracy---certification accuracy can both decrease and increase as the stride increases. 

For a few choices in striding, it is possible to not substantially change the accuracy of the ViT at classifying ablations, as shown in Figure \ref{appfig:stride-sweep}. 
For example, a ViT-B which normally obtains 38.3\% certified accuracy without striding, maintains certified accuracies of 37.6\% at stride $s=5$ and 36.8\% at stride $s=10$. For these small drops in certified accuracy, striding directly enables 5x or 10x faster inference times. 
% For example, ViT-B is $46.7\%$ accurate at classifying ablations with stride $s=10$---an order of magnitude faster than using all column ablations for only a $0.3\%$ drop in ablation accuracy. 

%% file: sections/app-block.tex
In this section, we investigate an alternative type of smoothing known as \textit{Block Smoothing}, previously investigated in the CIFAR-10 setting \citep{levine2020randomized}. In block smoothing, we ablate (square) blocks of pixels instead of columns of pixels. 
This procedure is prohibitively expensive for ImageNet due to its quadratic complexity. 
For example, smoothing a $224\times 224$ image with block ablations takes a majority vote over $224\times 224 = 50,176$ ablations, which is four orders of magnitude slower than a standard forward pass. 
We alleviate this obstacle for larger image settings such as ImageNet with the token-based speedups for ViTs from Section~\ref{sec:speed} and the striding from Section~\ref{sec:ablation_stride}. In combination, these improvements in speed allow us to perform a practical investigation into block smoothing on ImageNet. 
%to calculte certificates much faster than CNNs. This allows us to investigate in more detail the effect of different types of smoothing, specifically block smoothing.

\paragraph{Certification.}
Certification of derandomized smoothing models with block ablations is similar to that of column ablations, and depends on the maximum number of ablations in the ablation set that an adversarial patch can simultaneously intersect. 
Recall that for column ablations of size $b$, the certification threshold is $\Delta = m + b - 1$ ablations. 
For block ablations of size $b$ (where $b$ here is the side of the retained block/square of pixels), $\Delta=(m+b-1)^2$. 
The threshold can then be plugged as before into Equation~\eqref{eq:certify} to check whether the model is certifiably robust. 

\subsection{Practical inference speeds for block smoothing}
\begin{figure}[!htbp]
    \centering
    \includegraphics[width=.7\textwidth]{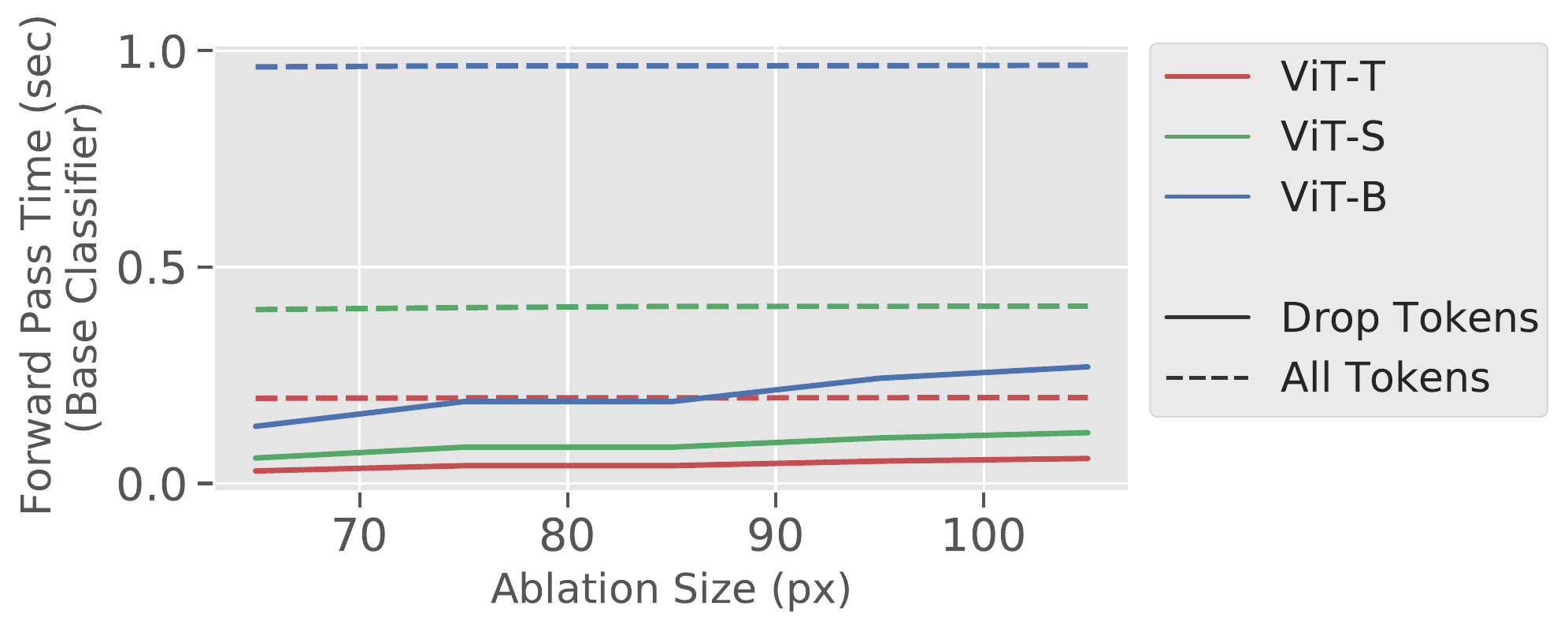}
    \caption{Average time to compute a forward pass for ViTs on 1024 block ablated images with varying ablation sizes with and without dropping masked tokens.}
    \label{fig:app-block-speed}
\end{figure}

We first demonstrate how dropping masked tokens significantly increases the speed of evaluating block ablations for the base classifier. In Figure~\ref{fig:app-block-speed}, we show that dropping masked tokens substantially reduces the time needed to process 1024 block ablations for various sizes of ViTs. This directly leads to a 4.85x speedup for ViT-S with ablation size 75. 

Even with this optimization, however, block smoothing is quite expensive.  
A forward pass through the smoothed model still requires around 50k passes through the base classifier. 
We thus leverage our second speedup from strided ablations and use \textit{strided} block smoothing. 
Similar to strided column ablations, for a stride length of $s$, we only consider block ablations that are $s$ pixels apart, vertically and horizontally. 
This changes the certification threshold $\Delta$ to be, $\Delta_{block+stride} = \lceil(m+s-1)/s\rceil^2$. With dropping fully masked tokens and using a stride of 10, a smoothed ViT-S using an ablation size of 75 is only 2.8x slower than a standard (non-robust) ResNet-50.

% \begin{figure}[!htbp]
%     \centering
%     \includegraphics[width=1\textwidth]{figures/block-smoothing/vary_all.pdf}
%     \caption{Strided block smoothing on ImageNet.}
%     \label{appfig:block-vary-patch-size}
% \end{figure}

\begin{figure}[!htbp]
    \centering
    \begin{subfigure}{1\textwidth}
        \includegraphics[width=1\textwidth]{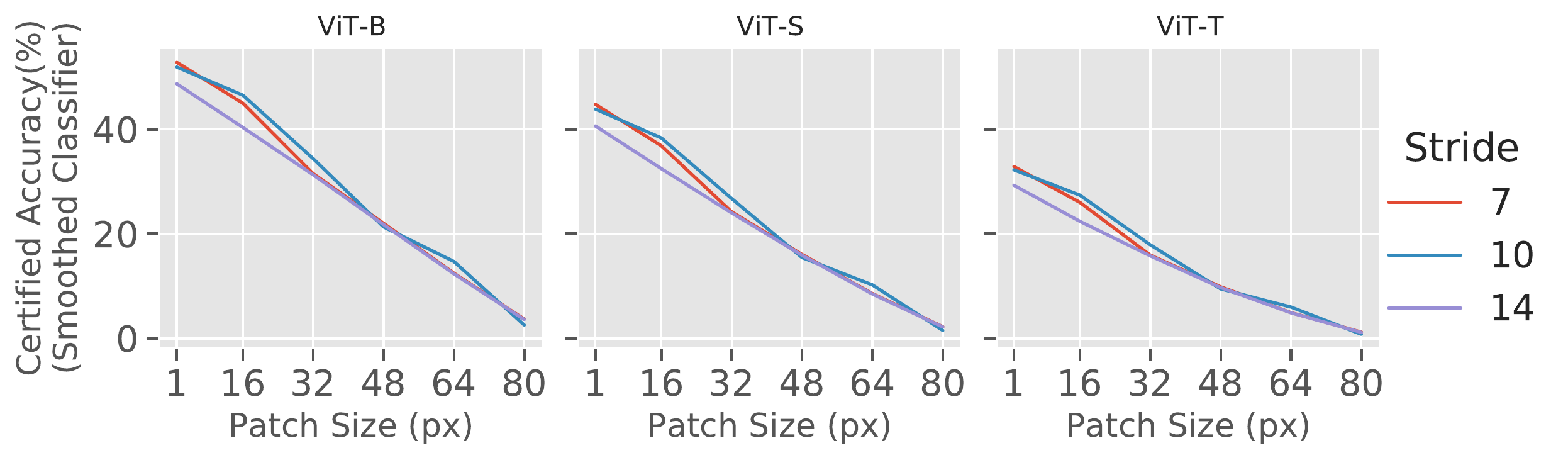}
        \caption{We fix the test-time ablation size at $b=75$ and plot the certified accuracy as a function of the adversarial patch size, for various stride length.}            
    \end{subfigure}
    \begin{subfigure}{1\textwidth}
        \includegraphics[width=1\textwidth]{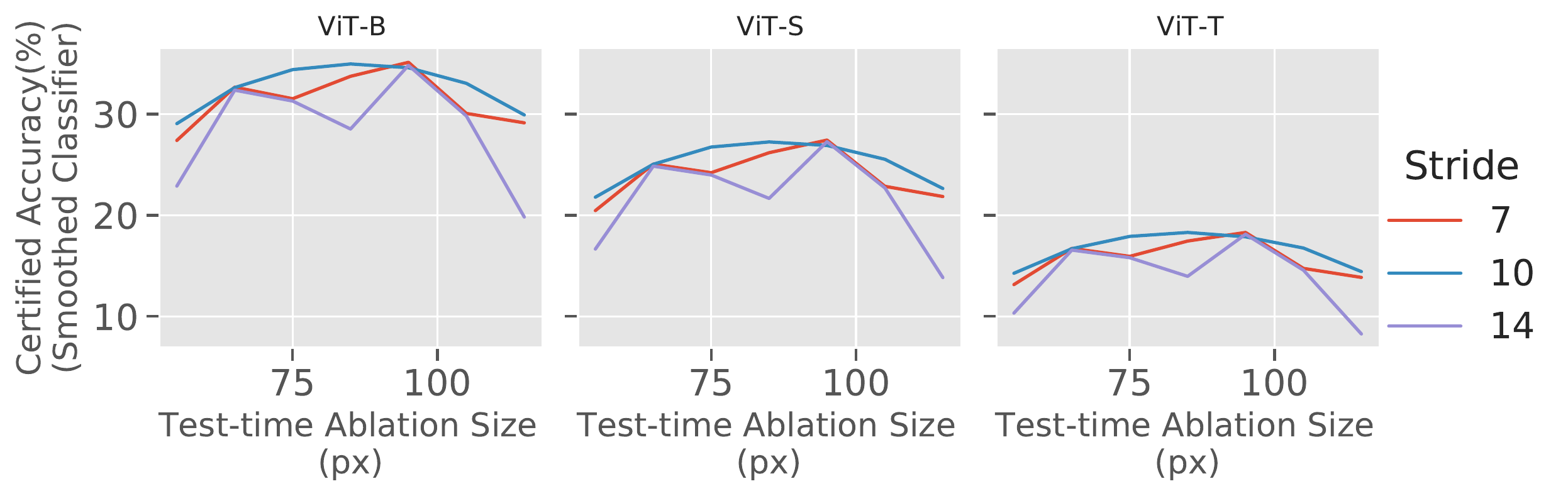}
        \caption{We fix the adversarial patch size $m=32$ and plot the certified accuracy as a function of the test-time ablation size, for various stride length.}            
    \end{subfigure}
    \caption{Strided block smoothing on ImageNet for a collection of ViT models trained with block ablations of size $b=75$.}
    \label{appfig:block-vary-ablation-size}
\end{figure}

\paragraph{Certified accuracy.}
We find that, despite an systematic search over stride length and block size (both at training and evaluation), block smoothing on ImageNet remains significantly worse than column smoothing. 
For example, with optimal stride and ablation size, we see up to 5\% lower certified accuracy than column smoothing on the largest model, ViT-B. 
We checked a range of ablation sizes from $55$ to $115$ as well as three stride lengths $\{7,10,14\}$ (Figure \ref{appfig:block-vary-ablation-size}).  

\begin{figure}[!htbp]
    \centering
    \includegraphics[width=.5\textwidth]{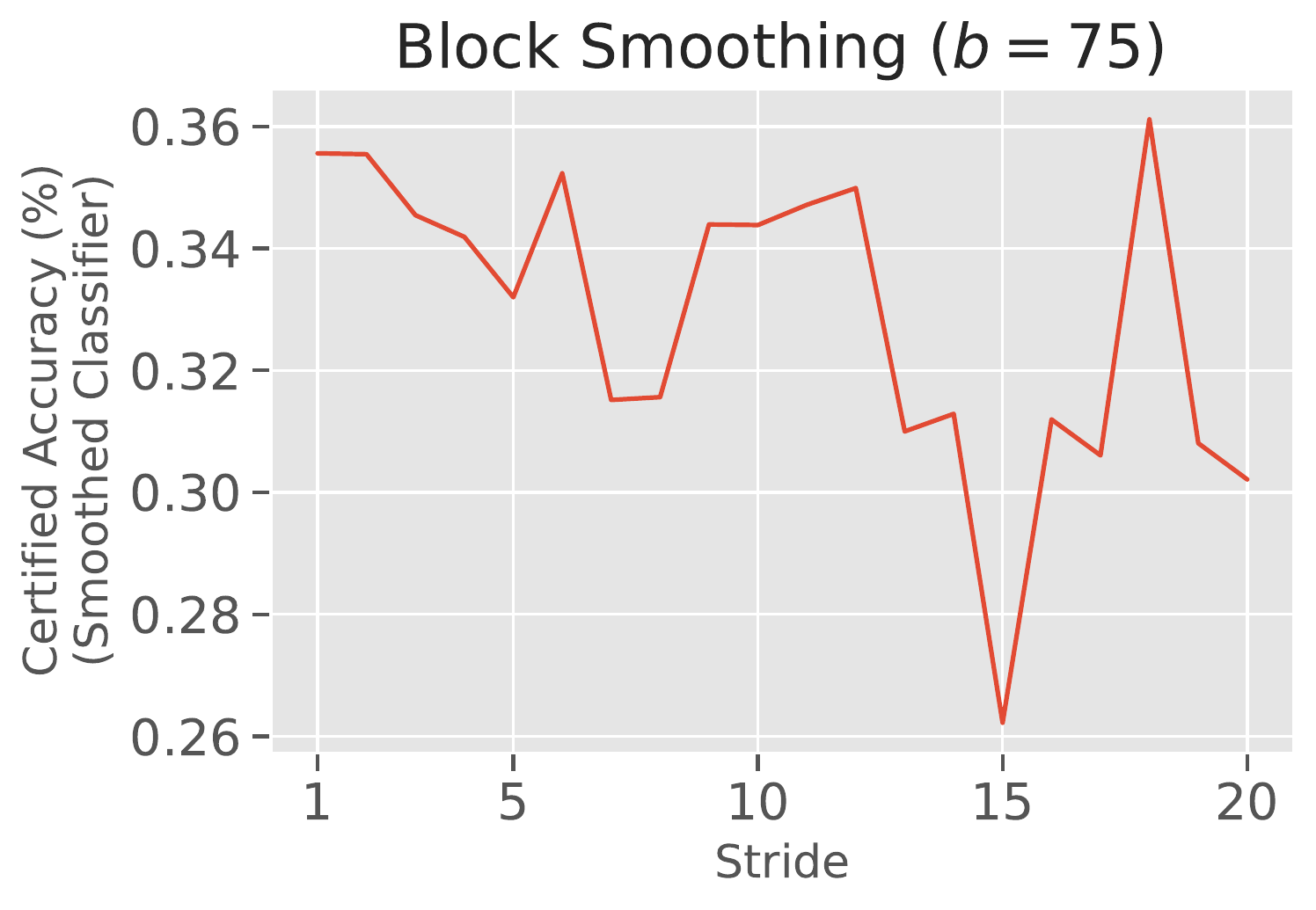}
    \caption{Strided block smoothing on ImageNet for ViT-B with a fixed ablation size $b=75$. The reported certified accuracy are against adversarial patches of size $32\times32$. Note how some stride lengths ($s=18$ for example) can achieve similar performance to non-strided block ablations ($s=1$).}
    \label{appfig:block-vary-stride}
\end{figure}

Similar to striding with column ablations, there is a significant amount of variability based on the stride length. 
To pinpoint the effect of striding, we certify one of the best-performing block sizes ($b=75$) over a full range of strides from $s=1$ to $s=20$  (Figure \ref{appfig:block-vary-stride}). 
This is a fairly expensive calculation, as using stride $s=1$ corresponds to the full block ablation with 50k ablations. 

Even when using all possible block ablations ($s=1$), block smoothing does not improve over column smoothing. 
However, we do find that certain stride lengths ($s=18$) can achieve similar performance to non-strided block ablations $(s=1)$, which means that we can speed up certification (by $18$x) without sacrificing certified accuracy. Thus, while our methods can make block smoothing computationally feasible, further investigation is needed to make block smoothing match column smoothing in terms of certified and standard accuracies.

% \begin{example}[Block Smoothing]
%     Block smoothing takes $\mathcal S_{b \times b}(x)$ to be the set of all column ablations, where an ablated image leaves a square block of $b \times b$ unmasked. Then, the maximum number of ablations that an $m\times m$ patch can intersect with is at most $\Delta=(m+b-1)^2$. 
% \end{example}

%% file: sections/tables.tex
\begin{table}[!htbp]
    \caption{\textbf{An extended version of Table~\ref{table:main_summary table}.} Summary of our ImageNet results and comparisons to certified patch defenses from the literature: Clipped Bagnet (CBG), Derandomized Smoothing (DS), and PatchGuard (PG). Time refers to the inference time for a batch of 1024 images, $b$ is the ablation size, and $s$ is the ablation stride. }
    \centering
    \begin{tabular}{l|cccc|r}
        \toprule
        \multicolumn{5}{c}{Standard and Certified Accuracy on ImageNet (\%) }\\
        \midrule
        Patch Size & Clean & 1\% pixels & 2\% pixels & 3\% pixels & Time (sec) \\
        \midrule
        CBN~\citep{zhang2020clipped} & 49.5 & 13.4 & 7.1&  3.1 & $3.05 \pm 0.02$\\ 
        DS~\citep{levine2020randomized} & 44.4 &17.7 & 14.0&  11.2 &  $149.52 \pm 0.33$\\ 
        % PG~\citep{xiang2021patchguard} & 55.1 &32.3 & 26.0&  19.7  & {}\\
        PG~\citep{xiang2021patchguard} (1\% pixels) & 55.1 &32.3  & 0.0&  0.0  &$3.05 \pm 0.02$\\
        PG~\citep{xiang2021patchguard} (2\% pixels) & 54.6 &26.0 & 26.0&  0.0  &$3.05 \pm 0.02$\\
        PG~\citep{xiang2021patchguard} (3\% pixels) & 54.1 &19.7 & 19.7&  19.7  &$3.05 \pm 0.02$\\
        \midrule
        \multicolumn{5}{l}{\textit{Vary Ablation Size (Stride = 1)}} \\ 
        \midrule					
        ResNet-18 (b = 19) & 50.6 & 24.1 & 19.8 & 16.9 & $39.84 \pm 0.97$\\
        ResNet-18 (b = 25) & 52.7 & 24.2 & 20.0 & 17.1 & $39.84 \pm 0.97$\\
        ResNet-18 (b = 37) & 54.3 & 22.4 & 18.6	& 15.7 & $39.84 \pm 0.97$\\
        ViT-T (b = 19) & 52.3 &    27.3 &  22.9 &  19.9  & $6.81 \pm 0.05$\\
        ViT-T (b = 25) & 53.7 &    26.9 &  22.8 &  19.7  & $6.82 \pm 0.05$\\
        ViT-T (b = 37) & 55.6 &    25.5 &  21.7 &  18.8  & $12.64 \pm 0.10$\\
        \midrule
        ResNet-50 (b = 19) & 51.5 & 22.8 & 18.3	& 15.3  & $149.52 \pm 0.33$\\
        ResNet-50 (b = 25) & 54.7 &	23.8 & 19.5	& 16.4  & $149.52 \pm 0.33$\\
        ResNet-50 (b = 37) & 57.8 & 23.1 & 	19.0 & 16.1  & $149.52 \pm 0.33$\\
        ViT-S (b = 19) & 63.5 &    36.8 &  31.6 &  27.9  & $14.00 \pm 0.16$\\
        ViT-S (b = 25) & 65.1 &    36.8 &  31.9 &  28.2  & $20.58 \pm 0.18$\\
        ViT-S (b = 37) & 67.1 &    35.3 &  30.7 &  27.1 & $20.61 \pm 0.16$\\
        \midrule
        WRN-101-2 (b = 19) & 61.4 &    33.3 &  28.1 &  24.1 & $694.50 \pm 0.58$\\
        WRN-101-2 (b = 25) & 64.2 &    34.3 &  29.1 &  25.3  & $694.50 \pm 0.58$\\
        WRN-101-2 (b = 37) & 67.2 &    33.7 &  28.8 &  25.2 & $694.50 \pm 0.58$\\
        ViT-B (b = 19) & 69.3 & 43.8 & 38.3	& 34.3 &  $31.51 \pm 0.17$\\
        ViT-B (b = 25) & 71.1 &    44.0 &  38.8 &  34.8 & $31.52 \pm 0.21$\\
        ViT-B (b = 37) & 73.2 &    43.0 &  38.2 &  34.1 & $58.74 \pm 0.17$\\
        \midrule
        \multicolumn{5}{l}{\textit{Vary Ablation Stride}} \\ 
        \midrule
        WRN-101-2 (b = 19, s = 5) & 61.1 &    30.1 &  27.3 &  21.9 & $138.90 \pm 0.12$\\
        WRN-101-2 (b = 19, s = 10) & 59.7 &    25.8 &  25.8 &  20.9 & $69.45 \pm 0.06$\\
        ViT-B (b = 19, s = 5) & 69.0 &    40.6 &  37.7 &  32.0 & $6.30 \pm 0.03$\\
        ViT-B (b = 19, s = 10) & 68.3 &    36.9 &  36.9 &  31.4 & $3.15 \pm 0.02$\\
        \midrule
        WRN-101-2 (b = 37, s = 5) & 66.9 &    32.6 &  27.2 &  24.7 & $138.90 \pm 0.12$\\
        WRN-101-2 (b = 37, s = 10) & 66.1 &    31.7 &  26.7 &  21.7 & $69.45 \pm 0.06$\\
        ViT-B (b = 37, s = 5) & 73.1 &    41.9 &  36.4 &  33.5 & $11.75 \pm 0.03$\\
        ViT-B (b = 37, s = 10) & 72.6 &    41.3 &  36.1 &  30.8 & $5.87 \pm 0.02$\\

        \bottomrule
    \end{tabular}
    \label{table:summary_table_imagenet}
\end{table}

\begin{table}[!htbp]
        \caption{\textbf{An extended version of Table~\ref{table:summary table_cifar}}. Summary of our CIFAR-10 results and comparisons to certified patch defenses from the literature: Clipped Bagnet (CBG), Derandomized Smoothing (DS), and PatchGuard (PG). $b$ is the column ablation size out of 32 pixels.}
        \centering
        \begin{tabular}{l|ccc}
            \toprule
            \multicolumn{4}{c}{Standard and Certified Accuracy on CIFAR-10 (\%) }\\
            \midrule
            Patch Size & Clean & $2\times2$ & $4\times4$ \\
            \midrule
            \multicolumn{4}{l}{\textit{Baselines} } \\
            \midrule
            CBN~\citep{zhang2020clipped} & 84.2 & 44.2 & 9.3\\ 
            DS~\citep{levine2020randomized} & 83.9 & 68.9 & 56.2\\ 
            PG~\citep{xiang2021patchguard} ($2\times 2$) & 84.7 &  69.2 & 0.0 \\
            PG~\citep{xiang2021patchguard} ($4\times 4$) & 84.6 &  57.7 & 57.7 \\
            \midrule
            \multicolumn{4}{l}{\textit{Smoothed models} } \\
            \midrule
            ResNet-18 (b = 4) & 83.6 &   67.0 &  54.2\\
            ViT-T (b = 4) & \textbf{85.5} &    \textbf{70.0} &  \textbf{58.5}  \\
            \midrule
            ResNet-50 (b = 4) & 86.4 &    71.6 &  59.0 \\
            ViT-S (b = 4) & \textbf{88.4} &    \textbf{75.0} &  \textbf{63.8} \\
            \midrule
            WRN-101-2 (b = 4) & 88.2 &    73.9 &  62.0 \\
            ViT-B (b = 4) & \textbf{90.8} &    \textbf{78.1} &  \textbf{67.6}  \\
            % \midrule
            \bottomrule
        \end{tabular}
        \label{table:summary_table_cifar_appendix}
\end{table}

\begin{table}[!htbp]
        \caption{Standard accuracies of regularly trained architectures vs. smoothed architectures with column ablations of size $b=4$ for CIFAR-10 and $b=19$ for ImageNet.}
        % \label{tab:standard}
        \begin{tabular}{llcccccc}
        \toprule
        && \multicolumn{6}{c}{Standard accuracy of architecture (\%)}\\
        \cmidrule(lr){3-8}
        {} &&  ViT-T &  ResNet-18 &  ViT-S &  ResNet-50 &  ViT-B &  WRN-101-2 \\
        \midrule
        \multirow{4}[-2]{*}{ImageNet} & Standard  &  72.03 &      69.76 &  79.72 &      76.13 &  81.74 &              78.85 \\
        & Smoothed  &  52.25 &      50.62 &  63.48 &      51.47 &  69.33 &              61.38 \\
        \cmidrule(lr){2-8}
        & Difference &  19.77 &      19.14 &  16.24 &      24.66 &  12.41 &              17.47 \\
        \midrule
        \multirow{4}[-2]{*}{CIFAR-10}& Standard   &  93.13 &      95.72 &  93.33 &      96.16 &  97.07 &              97.85 \\
        & Smoothed   &  85.53 &      88.41 &   86.39 &      83.57 &  90.75 &              88.20 \\
        \cmidrule(lr){2-8}
        & Difference &   7.60 &       7.31 &   6.94 &      12.59 &  6.32 &               9.65 \\
        \bottomrule
        \end{tabular}
\end{table}

\begin{table}[!htbp]
    \small
    \centering
    \caption{ImageNet certified models trained with ablations of size $19$, with a variety of test-time ablations sizes $b$ and ablation stride lengths $s$.}
    \def\arraystretch{0.95}
    \begin{tabular}{ccccccc}
    \toprule
            &    & {} & \multicolumn{1}{c}{Standard accuracy(\%)} & \multicolumn{3}{c}{Certified Accuracy(\%)} \\
    Architecture & s & b &           &      1\% pixels  & 2\% pixels  & 3\% pixels        \\
    \midrule
    \multirow{9}{*}{ResNet-18} & \multirow{3}{*}{1}  & 19 &       50.6 &    24.1 &  19.8 &  16.9 \\
            &    & 25 &       52.7 &    24.2 &  20.0 &  17.1 \\
            &    & 37 &       54.3 &    22.4 &  18.6 &  15.7 \\ \cmidrule{2-7}
            & \multirow{3}{*}{5}  & 19 &       50.3 &    21.5 &  19.3 &  15.3 \\
            &    & 25 &       52.4 &    22.1 &  17.9 &  15.8 \\
            &    & 37 &       54.2 &    21.5 &  17.4 &  15.4 \\ \cmidrule{2-7}
            & \multirow{3}{*}{10} & 19 &       49.3 &    18.7 &  18.7 &  14.8 \\
            &    & 25 &       51.5 &    21.5 &  17.3 &  13.6 \\
            &    & 37 &       53.7 &    21.0 &  17.1 &  13.5 \\ \cmidrule{1-7}
        %     \midrule
    \multirow{9}{*}{ViT-T} & \multirow{3}{*}{1}  & 19 &       52.3 &    \textbf{27.3} &  \textbf{22.9} &  \textbf{19.9} \\
            &    & 25 &       53.7 &    26.9 &  22.8 &  19.7 \\
            &    & 37 &       \textbf{55.6} &   25.5 &  21.7 &  18.8 \\ \cmidrule{2-7}
            & \multirow{3}{*}{5}  & 19 &       51.9 &    24.6 &  22.4 &  18.2 \\
            &    & 25 &       53.5 &    25.1 &  20.6 &  18.5 \\
            &    & 37 &       55.4 &    24.7 &  20.5 &  18.5 \\ \cmidrule{2-7}
            & 10 & 19 &       51.2 &    21.8 &  21.8 &  17.8 \\ 
            &    & 25 &       53.1 &    24.6 &  20.4 &  16.4 \\
            &    & 37 &       55.1 &    24.4 &  20.3 &  16.5 \\
            \midrule
            \midrule
        %     \midrule
    \multirow{9}{*}{ResNet-50} & \multirow{3}{*}{1}  & 19 &       51.5 &    22.8 &  18.3 &  15.3 \\
            &    & 25 &       54.7 &    23.8 &  19.5 &  16.4 \\
            &    & 37 &       57.8 &    23.1 &  19.0 &  16.1 \\ \cmidrule{2-7}
            & \multirow{3}{*}{5}  & 19 &       51.0 &    20.1 &  17.9 &  13.6 \\
            &    & 25 &       54.5 &    21.7 &  17.2 &  15.1 \\
            &    & 37 &       57.7 &    22.1 &  17.7 &  15.8 \\ \cmidrule{2-7}
            & \multirow{3}{*}{10} & 19 &       49.9 &    17.2 &  17.2 &  13.2 \\
            &    & 25 &       53.7 &    21.0 &  16.7 &  12.8 \\
            &    & 37 &       57.1 &    21.7 &  17.6 &  13.7 \\ \cmidrule{1-7}
        %     \midrule
    \multirow{9}{*}{ViT-S} & \multirow{3}{*}{1}  & 19 &       63.5 &    \textbf{36.8} &  31.6 &  27.9 \\
            &    & 25 &       65.1 &    \textbf{36.8} &  \textbf{31.9} &  \textbf{28.2} \\
            &    & 37 &       \textbf{67.1} &    35.3 &  30.7 &  27.1 \\ \cmidrule{2-7}
            & \multirow{3}{*}{5}  & 19 &       63.1 &    33.8 &  31.1 &  25.7 \\
            &    & 25 &       64.9 &    34.4 &  29.3 &  26.7 \\
            &    & 37 &       67.0 &    34.3 &  29.1 &  26.7 \\ \cmidrule{2-7}
            & \multirow{3}{*}{10} & 19 &       62.2 &    30.3 &  30.3 &  25.2 \\
            &    & 25 &       64.3 &    33.9 &  28.7 &  23.7 \\
            &    & 37 &       66.5 &    33.8 &  29.0 &  24.2 \\
            \midrule
            \midrule
    \multirow{9}{*}{WRN-101} & \multirow{3}{*}{1}  & 19 &       61.4 &    33.3 &  28.1 &  24.1 \\
            &    & 25 &       64.2 &    34.3 &  29.1 &  25.3 \\
            &    & 37 &       67.2 &    33.7 &  28.8 &  25.2 \\ \cmidrule{2-7}
            & \multirow{3}{*}{5}  & 19 &       61.1 &    30.1 &  27.3 &  21.9 \\
            &    & 25 &       63.8 &    31.8 &  26.3 &  23.7 \\
            &    & 37 &       66.9 &    32.6 &  27.2 &  24.7 \\ \cmidrule{2-7}
            & \multirow{3}{*}{10} & 19 &       59.7 &    25.8 &  25.8 &  20.9 \\
            &    & 25 &       62.7 &    30.5 &  25.3 &  20.5 \\
            &    & 37 &       66.1 &    31.7 &  26.7 &  21.7 \\ \cmidrule{1-7}
        %     \midrule
    \multirow{9}{*}{ViT-B} & \multirow{3}{*}{1}  & 19 &       69.3 &    43.8 &  38.3 &  34.3 \\
            &    & 25 &       71.1 &    \textbf{44.0} &  \textbf{38.8} &  \textbf{34.8} \\
            &    & 37 &       \textbf{73.2} &    43.0 &  38.2 &  34.1 \\ \cmidrule{2-7}
            & \multirow{3}{*}{5}  & 19 &       69.0 &    40.6 &  37.7 &  32.0 \\
            &    & 25 &       70.8 &    41.6 &  36.0 &  33.0 \\
            &    & 37 &       73.1 &    41.9 &  36.4 &  33.5 \\ \cmidrule{2-7}
            & \multirow{3}{*}{10} & 19 &       68.3 &    36.9 &  36.9 &  31.4 \\
            &    & 25 &       70.3 &    40.9 &  35.2 &  29.8 \\
            &    & 37 &       72.6 &    41.3 &  36.1 &  30.8 \\
    \bottomrule
    \end{tabular}
    \def\arraystretch{1}
\end{table}